\PassOptionsToPackage{pagebackref=false,breaklinks=true,bookmarks=false,hyperfootnotes=false}{hyperref}
\documentclass[a4paper,fleqn]{cas-dc}

\usepackage[authoryear,round]{natbib}

\definecolor{rblue}{rgb}{0,0.5,1}
\definecolor{hollywoodcerise}{rgb}{0.96, 0.0, 0.63}
\definecolor{lasallegreen}{rgb}{0.03, 0.47, 0.19}
\definecolor{hanpurple}{rgb}{0.32, 0.09, 0.98}
\definecolor{green(pigment)}{rgb}{0.0, 0.65, 0.31}

\usepackage{subcaption}

\usepackage{textcomp}
\usepackage{gensymb}
\usepackage{pifont}
\usepackage{threeparttable}

\def\ie{\emph{i.e.}}

\definecolor{mygray}{gray}{.9}
\definecolor{myred}{RGB}{192, 0, 0}
\definecolor{mygreen}{RGB}{0, 176, 80}

\definecolor{colorVehicle}{RGB}{100, 150, 245}
\definecolor{colorPedestrian}{RGB}{255, 30, 30}
\definecolor{colorRoad}{RGB}{255, 0, 255}
\definecolor{colorBuilding}{RGB}{255, 200, 0}
\definecolor{colorVegetation}{RGB}{0, 175, 0}
\definecolor{colorTerrain}{RGB}{150, 240, 80}
\makeatletter
\newcommand{\vehicle@quadoccfreq}{0.90}
\newcommand{\pedestrian@quadoccfreq}{2.09} 
\newcommand{\road@quadoccfreq}{52.34}
\newcommand{\building@quadoccfreq}{14.00}
\newcommand{\vegetation@quadoccfreq}{22.82}
\newcommand{\terrain@quadoccfreq}{7.85}
\newcommand{\quadoccfreq}[1]{{\csname #1@quadoccfreq\endcsname}}
\makeatother

\definecolor{hthreeoRoad}{RGB}{128,64,128}        
\definecolor{hthreeoSidewalk}{RGB}{244,35,232}    
\definecolor{hthreeoBuilding}{RGB}{70,70,70}      
\definecolor{hthreeoVegetation}{RGB}{107,142,35}  
\definecolor{hthreeoCar}{RGB}{0,0,142}            
\definecolor{hthreeoTruck}{RGB}{0,0,70}           
\definecolor{hthreeoBus}{RGB}{0,60,100}           
\definecolor{hthreeoTwoWheeler}{RGB}{0,0,230}     
\definecolor{hthreeoPerson}{RGB}{220,20,60}       
\definecolor{hthreeoPole}{RGB}{153,153,153}       

\makeatletter
\newcommand{\road@hthreeofreqHomo}{34.04}
\newcommand{\sidewalk@hthreeofreqHomo}{26.24}
\newcommand{\building@hthreeofreqHomo}{10.00}
\newcommand{\vegetation@hthreeofreqHomo}{24.02}
\newcommand{\car@hthreeofreqHomo}{2.92}
\newcommand{\truck@hthreeofreqHomo}{0.99}
\newcommand{\bus@hthreeofreqHomo}{0.25}
\newcommand{\twoWheeler@hthreeofreqHomo}{0.07}
\newcommand{\person@hthreeofreqHomo}{0.97}
\newcommand{\pole@hthreeofreqHomo}{0.50}
\newcommand{\hthreeofreqHomo}[1]{{\csname #1@hthreeofreqHomo\endcsname}}

\newcommand{\road@hthreeofreqHeter}{34.04}
\newcommand{\sidewalk@hthreeofreqHeter}{26.24}
\newcommand{\building@hthreeofreqHeter}{10.00}
\newcommand{\vegetation@hthreeofreqHeter}{24.02}
\newcommand{\car@hthreeofreqHeter}{2.92}
\newcommand{\truck@hthreeofreqHeter}{0.99}
\newcommand{\bus@hthreeofreqHeter}{0.25}
\newcommand{\twoWheeler@hthreeofreqHeter}{0.07}
\newcommand{\person@hthreeofreqHeter}{0.97}
\newcommand{\pole@hthreeofreqHeter}{0.50}
\newcommand{\hthreeofreqHeter}[1]{{\csname #1@hthreeofreqHeter\endcsname}}
\makeatother

\newcommand{\ColHeadHomo}[3]{\rotatebox{90}{\textcolor{#1}{$\blacksquare$} \makecell[l]{#2 \\ (\hthreeofreqHomo{#3}\%)}}}
\newcommand{\ColHeadHeter}[3]{\rotatebox{90}{\textcolor{#1}{$\blacksquare$} \makecell[l]{#2 \\ (\hthreeofreqHeter{#3}\%)}}}

\definecolor{r1}{RGB}{193, 58, 33}
\definecolor{r2}{rgb}{0.55, 0.0, 0.55}
\definecolor{r3}{RGB}{0, 110, 184}

\definecolor{codegreen}{rgb}{0.0,0.6,0.0}

\definecolor{ceiling}{RGB}{214,  38, 40}   %
\definecolor{floor}{RGB}{43, 160, 4}     %
\definecolor{wall}{RGB}{158, 216, 229}  %
\definecolor{window}{RGB}{114, 158, 206}  %
\definecolor{chair}{RGB}{204, 204, 91}   %
\definecolor{bed}{RGB}{255, 186, 119}  %
\definecolor{sofa}{RGB}{147, 102, 188}  %
\definecolor{table}{RGB}{30, 119, 181}   %
\definecolor{tvs}{RGB}{160, 188, 33}   %
\definecolor{furniture}{RGB}{255, 127, 12}  %
\definecolor{other}{RGB}{196, 175, 214} 

\makeatletter
\newcommand{\ceiling@nyufreq}{1.37}
\newcommand{\floor@nyufreq}{17.58}
\newcommand{\wall@nyufreq}{15.26}
\newcommand{\window@nyufreq}{1.99}
\newcommand{\chair@nyufreq}{3.01}
\newcommand{\bed@nyufreq}{7.08}
\newcommand{\sofa@nyufreq}{4.70}
\newcommand{\table@nyufreq}{4.31}
\newcommand{\tvs@nyufreq}{0.47}
\newcommand{\furniture@nyufreq}{30.04}
\newcommand{\other@nyufreq}{14.19}
\newcommand{\nyufreq}[1]{{\csname #1@nyufreq\endcsname}}
\makeatother

\newcommand{\ColHeadNyu}[3]{\rotatebox{90}{\textcolor{#1}{$\blacksquare$} \makecell[l]{#2 \\ (\nyufreq{#3}\%)}}}

\hypersetup{colorlinks,linkcolor={red},citecolor={hanpurple},urlcolor={magenta},pdfauthor={Anonymous}}
\newif\ifshowauthors
\showauthorstrue

\begin{document}

\shorttitle{O3N: Omnidirectional Open-Vocabulary Occupancy Prediction}

\ifshowauthors
    \shortauthors{Duan et~al.}
\else
    \shortauthors{Anonymous}
\fi

\title[mode=title]{O3N: Omnidirectional Open-Vocabulary Occupancy Prediction}

\tnotetext[1]{This work was supported in part by the National Natural Science Foundation of China (Grant No. 62473139), in part by the Hunan Provincial Research and Development Project (Grant No. 2025QK3019), and in part by the State Key Laboratory of Autonomous Intelligent Unmanned Systems (the opening project number ZZKF2025-2-10).}

\ifshowauthors

\author[1,2]{Mengfei Duan}

\fnmark[1]

\author[3,4]{Hao Shi}
\fnmark[1]

\author[1,2]{Fei Teng}
\author[1,2]{Guoqiang Zhao}
\author[1,2]{Yuheng Zhang}
\author[1,2]{Zhiyong Li}
\cormark[1]
\ead{zhiyong.li@hnu.edu.cn}
\author[1,2]{Kailun Yang}
\cormark[1]
\ead{kailun.yang@hnu.edu.cn}

\affiliation[1]{organization={School of Artificial Intelligence and Robotics, Hunan University},
  city={Changsha}, country={China}}
\affiliation[2]{organization={National Engineering Research Center of Robot Visual Perception and Control Technology, Hunan University},
  city={Changsha}, country={China}}
\affiliation[3]{organization={Ant Group}, city={Hangzhou}, country={China}}
\affiliation[4]{organization={State Key Laboratory of Extreme Photonics and Instrumentation, Zhejiang University},
  city={Hangzhou}, country={China}}

\cortext[1]{Corresponding authors: Kailun Yang and Zhiyong Li.}

\fntext[1]{Equal contribution.}

\fi

\begin{abstract}
Understanding and reconstructing the 3D world through omnidirectional perception is becoming increasingly important for autonomous agents and embodied systems operating in complex and dynamic environments. However, existing vision-based 3D occupancy prediction methods are constrained by limited perspective inputs and a predefined training distribution, making them difficult to support intelligent systems that require comprehensive and safe perception of scenes in open-world exploration. To address this, we present O3N, the first framework for open-vocabulary occupancy prediction from a single omnidirectional RGB image. O3N embeds omnidirectional voxels in a polar-spiral topology via the Polar-spiral Mamba (PsM) module, enabling continuous spatial representation and long-range context modeling across 360{\textdegree}. The Occupancy Cost Aggregation (OCA) module introduces a principled mechanism for unifying geometric and semantic supervision within the voxel space, ensuring consistency between reconstructed geometry and underlying semantic structure. Moreover, Natural Modality Alignment (NMA) establishes a gradient-free alignment pathway that harmonizes visual features, voxel embeddings, and text semantics, forming a consistent \textit{``pixel-voxel-text''} representation triad for open-world perception. Extensive experiments on multiple models demonstrate that our method not only achieves state-of-the-art performance on QuadOcc and Human360Occ benchmarks but also exhibits remarkable cross-scene generalization and semantic scalability, highlighting its potential to support scalable open-world 3D perception. 
\ifshowauthors
The source code will be made publicly available at \url{https://github.com/MengfeiD/O3N}.
\else
The source code will be made publicly available.
\fi
\end{abstract}


\maketitle

\begin{figure*}[pos=t]
  \centering
  \includegraphics[width=0.98\textwidth]{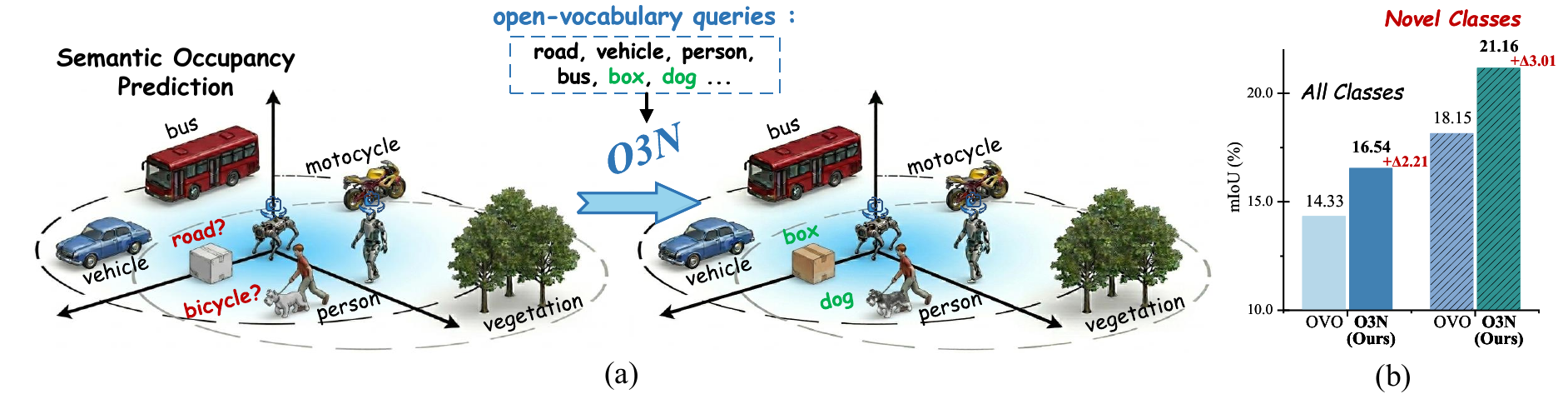}
  \caption{
        (a) \textbf{Omnidirectional Open-Vocabulary Occupancy Prediction.}
        Given a single omnidirectional RGB image as visual input, together with textual category queries, O3N predicts semantic occupancy of categories beyond the training vocabulary. 
        In this conceptual example, a closed-set semantic occupancy prediction model may incorrectly classify a \emph{\textbf{\textcolor{mygreen}{box}}} as \emph{\textbf{\textcolor{myred}{road}}}, or a \emph{\textbf{\textcolor{mygreen}{dog}}} as a \emph{\textbf{\textcolor{myred}{bicycle}}}. 
        By providing the corresponding category names as textual queries, O3N enables open-world semantic understanding of previously unseen objects.
        (b) \textbf{Results on the QuadOcc~\citep{shi2026oneocc} benchmark.} O3N reaches $16.54$ mIoU and $21.16$ \emph{Novel} mIoU, respectively, achieving state-of-the-art performance.
  }
  \label{fig:teaser}
\end{figure*}

\section{Introduction}
\label{sec:intro}

Comprehensive 3D scene understanding is fundamental to intelligent perception systems operating in complex and unstructured environments, facilitating downstream applications such as robotic navigation, autonomous exploration, and spatial reasoning~\citep{lu2025gsff,lin2026stg}.
Such systems require visual perception that provides comprehensive scene coverage, accurate geometric reasoning, and generalizable semantic understanding~\citep{wu2024embodiedocc,sun2025dynamic_embodied_occupancy,shi2026oneocc,cao2026occany}. 
Omnidirectional images provide broader spatial coverage and richer contextual information to fulfill this objective~\citep{gao2022review,zheng2025one_flight,zhu2026panoramic_scene_analysis,ai2025survey}. 
In parallel, vision-based 3D semantic occupancy prediction~\citep{wang2023openoccupancy,tian2023occ3d,wang2024multi,guo2025overlapocc,tian2026mambaocc} has emerged as a promising approach to jointly characterize spatial geometry and semantic information by lifting 2D visual observations into structured 3D voxel representations~\citep{cao2022monoscene,huang2023tpvformer,ma2024cotr,wei2023surroundocc,zuo2025quadricformer}.
Despite this, these methods assume that scene understanding is the recognition of a limited set of labels, which typically limits the models to predefined categories, causing them to struggle with complex and dynamic objects in open-world environments~\citep{zheng2024veon,jiang2024openocc,boeder2025langocc}. 
Thus, open-vocabulary understanding is crucial, as it enables the identification and prediction of unknown object categories without predefined category sets~\citep{li2022language,cho2024cat_seg,xie2024sed,tai2026open_sam3d}, facilitating flexible and scalable semantic representation.

\begin{figure*}[pos=!t]
    \centering
    \includegraphics[width=\textwidth]{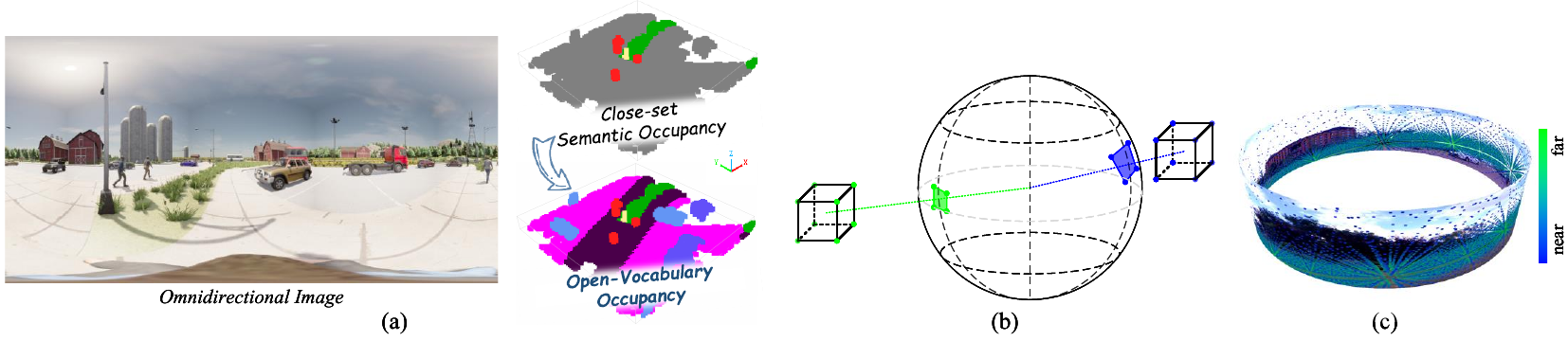}
    \caption{
    \textbf{Omnidirectional open-vocabulary occupancy prediction and its inherent imaging challenges.}
    (a) Omnidirectional open-vocabulary occupancy prediction aims to reconstruct the 3D geometric and semantic structure of a scene from a single omnidirectional RGB image;
    (b) The spatial distribution of visual evidence is affected by distance-dependent perspective scaling and latitude-dependent distortion introduced by the equirectangular projection (ERP); 
    (c) When uniformly spaced Cartesian voxel centers are projected onto an omnidirectional image, distant voxels tend to share densely concentrated image samples, resulting in non-uniform pixel-to-voxel correspondences.
    }
    \label{fig:vox2ODIs}
\end{figure*}

To the best of our knowledge, as illustrated in Fig.~\ref{fig:teaser}, this is the first work to investigate omnidirectional open-vocabulary occupancy prediction, unifying omnidirectional visual perception with 3D geometric-semantic prediction under open-ended semantics, thereby providing a more comprehensive scene understanding (see Fig.~\ref{fig:vox2ODIs}a). 
Inevitably, the challenges posed by panoramic imaging remain significant, including severe geometric distortions~\citep{zhang2024behind} and non-uniform sampling~\citep{zheng2025one_flight} introduced by the Equirectangular Projection (ERP). These issues limit the precise perception of spatial geometry and the effective management of high-dimensional open-vocabulary semantics. 
As illustrated in Fig.~\ref{fig:vox2ODIs}b, distance-dependent scale variations cause scene structures of comparable physical sizes to occupy progressively smaller angular regions as their distance from the camera increases. In addition, latitude-dependent distortion introduced by ERP further contributes to the non-uniform distribution of visual evidence. 
Consequently, multiple distant voxel centers may be projected onto densely concentrated or overlapping image regions, resulting in imbalanced pixel-to-voxel correspondences and insufficiently discriminative visual evidence for voxel-level semantic prediction. 
This characteristic amplifies the risk of overfitting~\citep{tan2023ovo,vobecky2023pop_3d} in the \textit{``pixel-voxel-text''} triplet feature alignment strategy, which relies on partially visible semantics under uneven data distribution, thereby leading to misalignment of novel semantics in the joint embedding space.

To address these challenges, we propose O3N, the first framework for open-vocabulary 3D occupancy prediction from a single omnidirectional RGB image. 
The novelty is threefold: 
(1) For intelligent systems operating in complex and unstructured environments, omnidirectional perception provides comprehensive spatial observations, enabling more complete 3D geometric modeling and semantic understanding of surrounding scenes. 
Cylindrical voxel partitioning provides a natural representation for omnidirectional geometry, but introduces spatial discontinuities and non-uniform geometric organization that can disrupt continuous $360^\circ$ scene modeling. We therefore design a dual-branch Polar-spiral Mamba (PsM) module that jointly models cylindrical and Cartesian voxel representations. 
By traversing polar features in a spiral order and progressively integrating them with Cartesian voxel features, PsM captures fine-grained geometric structures and long-range semantic dependencies while alleviating discontinuities near the angular boundaries of cylindrical representations~\citep{zuo2023pointocc,ming2025occcylindrical,wu2025omniocc}.
(2) The non-uniform distribution of panoramic visual evidence leads to imbalanced voxel sampling, making independent voxel-wise semantic matching less reliable. 
Prior studies in 2D open-vocabulary segmentation~\citep{cho2024cat_seg,xie2024sed} show that refining the matching costs between text and pixel features via cost aggregation is beneficial for semantic learning. 
Motivated by this observation, we propose Occupancy Cost Aggregation (OCA), which constructs an occupancy cost volume and aggregates it through spatial and class-wise reasoning, yielding more spatially coherent and semantically discriminative matching. This structured reasoning improves geometry–semantic consistency across the surrounding space. 

(3) Moreover, the inherent modality gaps among text, pixel, and voxel embeddings can introduce noisy cross-modal correspondences and weaken generalization to unseen semantics~\citep{tan2023ovo,cho2024cat_seg,zhao2025dpseg}. 
To reduce such noisy correspondences and mitigate overfitting to the limited semantic categories observed during training, we propose a gradient-free aggregation mechanism termed Natural Modality Alignment (NMA). 
NMA harmonizes these heterogeneous representations without introducing additional gradient-based alignment optimization, thereby improving semantic transfer to previously unseen categories and enhancing the extensibility of perception systems in open-world mobility scenarios.
Extensive experiments on the challenging QuadOcc and Human360Occ benchmarks~\citep{shi2026oneocc} demonstrate that O3N achieves state-of-the-art open-vocabulary occupancy prediction, achieving ${+}2.21$ mIoU and ${+}3.01$ \emph{Novel} mIoU on QuadOcc and ${+}0.86$ and ${+}1.54$ on Human360Occ over the baseline.
These consistent improvements, particularly in recognizing novel semantics, highlight the potential of O3N to support scalable and generalizable 3D perception in complex open-world environments.

At a glance, we deliver the following contributions:
\begin{itemize}
    \item We introduce the omnidirectional open-vocabulary occupancy prediction task and propose O3N, the first framework to perform this task from a single omnidirectional RGB image.
    \item To fully adapt to panoramic geometry and the density distribution of visual semantic information, the Polar-spiral Mamba (PsM) module, Occupancy Cost Aggregation (OCA), and gradient-free Natural Modality Alignment (NMA) are introduced in O3N for enhanced spatial-semantic modeling.
    \item Extensive experiments on the QuadOcc and Human360Occ benchmarks with multiple occupancy backbones demonstrate that O3N consistently outperforms the baseline and achieves competitive performance against several fully supervised methods, verifying its effectiveness and generalizability.
\end{itemize}

\section{Related Work}
\label{sec:relatedWork}

\subsection{Panoramic Semantic Scene Understanding}  
Omnidirectional and surround-view perception has attracted increasing attention in intelligent systems due to its capability of providing comprehensive environmental awareness beyond the limited field of view of conventional cameras. 
Such perception paradigms have been widely explored in applications including intelligent transportation systems, autonomous robots, and embodied intelligent robotics, where continuous observation of surrounding environments is essential for safe interaction and adaptive operation.
However, panoramic understanding is challenged by distortions and uneven sampling~\citep{zheng2025one_flight,hu2022distortion,li2023sgat4pass,li2026rel_sf4pass} in omnidirectional images.
Early panoramic semantic segmentation methods decomposed panoramic images into multiple perspective patches and processed them independently~\citep{yang2019pass,yang2020ds}.
Subsequent works reduced the dependence on panoramic annotations through domain adaptation~\citep{zhang2024behind,ma2021densepass,zheng2023both,zhang2022bending,zheng2023look_neighbor}, source-free adaptation~\citep{zheng2024semantics,zheng2024360sfuda++,chang2026denoise}, and pseudo-label generation~\citep{zhang2024goodsam,zhong2025omnisam}.
Open-vocabulary panoramic segmentation~\citep{zheng2024ops} has also been explored through distortion-aware augmentation and contrastive alignment~\citep{jiang2026augmenting}.
In parallel, dedicated architectures model panoramic distortion and long-range dependencies~\citep{orhan2022semantic,xu2025mamba4pass,cao2024geometric,lan2025deformable,guttikonda2024single,yuan2023laformer,jiang2025gaussian}, while other studies investigate self-supervised learning~\citep{tan2025dasc} and knowledge distillation~\citep{kim2022pasts}.

Beyond image-plane segmentation, panoramic semantic mapping projects omnidirectional visual observations into Bird's-Eye-View (BEV) representations.
Existing methods adopt either surround-view fisheye images~\citep{samani2023f2bev,yogamani2024fisheyebevseg,liu2025articubevseg}, a single panoramic image~\citep{wenke2025dur360bev,wei2024onebev,teng2024360bev} as input, or employ cross-modal knowledge distillation~\citep{sun2026kd360}.
More recently, panoramic scene understanding has been extended to 3D semantic occupancy prediction.
Existing approaches exploit different sensing configurations, including multiple fisheye cameras~\citep{pan2024generocc,yang2026parkocc}, depth information~\citep{wu2025omniocc}, and point clouds~\citep{zheng2025doracamom}, to enhance geometric reconstruction and semantic reasoning in 3D space.
OneOcc~\citep{shi2026oneocc} further investigates camera-only panoramic semantic occupancy prediction for legged robots.
Nevertheless, these methods are typically trained under closed-set supervision and restrict their predictions to predefined semantic categories, limiting their ability to recognize unseen semantics in open-world environments.

\subsection{Open-vocabulary Occupancy Prediction} 
A dominant line of work in 3D occupancy prediction follows a closed-set paradigm, where models are trained to recognize predefined semantic categories from LiDAR~\citep{yang2025daocc} or multi-view inputs~\citep{cao2022monoscene,wei2023surroundocc,li2023voxformer,tong2023scene_as_occupancy,huang2024selfocc,shi2024occupancy_set_points,zhu2024nucraft}. 
These approaches focus on efficient spatial representation~\citep{huang2023tpvformer,ma2024cotr,zhang2023occformer,tang2024sparseocc,huang2024gaussianformer}, contextual reasoning and semantic generalization~\citep{oh2025_3d_prototype,duan2025sdgocc,zhang2025occloff}, yet remain limited to fixed vocabularies. Open-vocabulary occupancy prediction methods enable models to predict the occupancy of new, unseen object categories from natural language input without explicit labels. 
OVO~\citep{tan2023ovo} pioneered a framework for open-vocabulary occupancy by distilling knowledge from a frozen 2D open-vocabulary segmenter and CLIP text encoder~\citep{radford2021learning} into a 3D model. 
Subsequent studies have explored multimodal fusion~\citep{jiang2024openocc}, self-supervised learning~\citep{vobecky2023pop_3d,gao2025loc}, foundation-model adaptation and training-free inference~\citep{zheng2024veon,zhou2025autoocc,jiang2026freeocc}, neural radiance field ensembles~\citep{boeder2025langocc}, Gaussian-based scene representations~\citep{li2025pgocc,zhou2026monocular_open_vocabulary,zhao2026shelfgaussian}, and pseudo-labeled reconstruction~\citep{yu2024language,li2025ago} to advance open-vocabulary 3D scene perception.
Unlike existing methods that rely on fixed input perspectives and predefined training categories, our method leverages flexible omnidirectional vision and open-vocabulary understanding to break such rigid limitations, enabling more robust, generalizable scene comprehension beyond traditional closed-world assumptions.

\begin{figure*}[pos=!t]
    \centering
    \includegraphics[width=\textwidth]{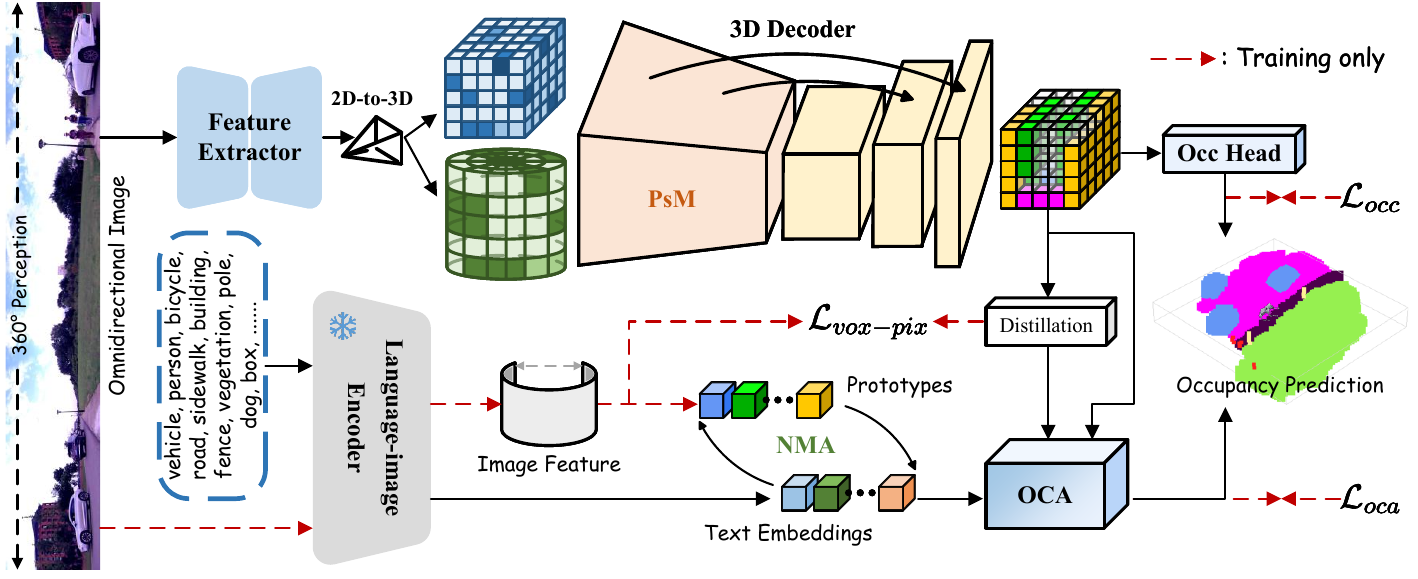}
    \caption{
    \textbf{O3N architecture.} O3N takes a single equirectangular omnidirectional image as visual input and is fully end-to-end trained. The image feature and text embeddings are pre-extracted via the language-image encoder. The 3D decoder, enhanced with the proposed Polar-spiral Mamba (PsM) module, captures both geometric and semantic dependencies across directions. The Occupancy Cost Aggregation (OCA) and Natural Modality Alignment (NMA) modules integrate pixel, voxel, and text modalities to achieve consistent open-vocabulary semantic reasoning in 3D space. We efficiently train the O3N through semantic occupancy of visible classes, occupancy cost aggregation, and voxel-pixel consistency during training.
    }
    \label{fig:overview}
\end{figure*}

\section{Methodology}
\label{sec:methodology}
An overview of the proposed O3N framework is depicted in Fig.~\ref{fig:overview}. 
In this work, we investigate open-vocabulary 3D occupancy prediction from a single omnidirectional RGB image. 
The omnidirectional input provides full-view spatial coverage and continuous contextual cues, offering a more complete visual basis for geometric and semantic reasoning in 3D space. 
Unlike prior approaches that rely on multimodal inputs or multi-stage pipelines~\citep{zheng2024veon,jiang2024openocc,vobecky2023pop_3d,gao2025loc,yu2024language}, O3N is, to the best of our knowledge, the first framework developed for this setting, providing a unified solution for comprehensive scene understanding.

Specifically, O3N includes the following components: (1) a visual feature extractor for the omnidirectional RGB image; (2) a 2D-to-3D view transformation that generates both 3D cube and cylindrical voxel representations; (3) a 3D decoder capable of learning fine-grained spatial geometry and semantic information; and (4) an occupancy prediction head. Through the close integration of these modules, O3N can not only perform closed-set semantic occupancy prediction but also demonstrate strong transferability. 
Furthermore, the proposed Polar-spiral Mamba (PsM, see Sec.~\ref{sec:polar_spiral_mamba}) module enhances omnidirectional spatial perception capabilities effectively. 
For 3D open-vocabulary perception, the Occupancy Cost Aggregation (OCA, Sec.~\ref{sec:occupancy_cost_aggregation}) and a gradient-free Natural Modality Alignment (NMA, Sec.~\ref{sec:natural_modality_alignment}) technique are proposed to enforce the \textit{``pixel-voxel-text''} semantic consistency and effectively mitigate overfitting. 
This unified design not only enhances perception completeness but also endows 3D occupancy prediction with semantic plasticity: a key step toward scalable world modeling.

\subsection{Polar-spiral Mamba Module}
\label{sec:polar_spiral_mamba}

\begin{figure}[pos=!t]
    \centering
    \includegraphics[width=\linewidth]{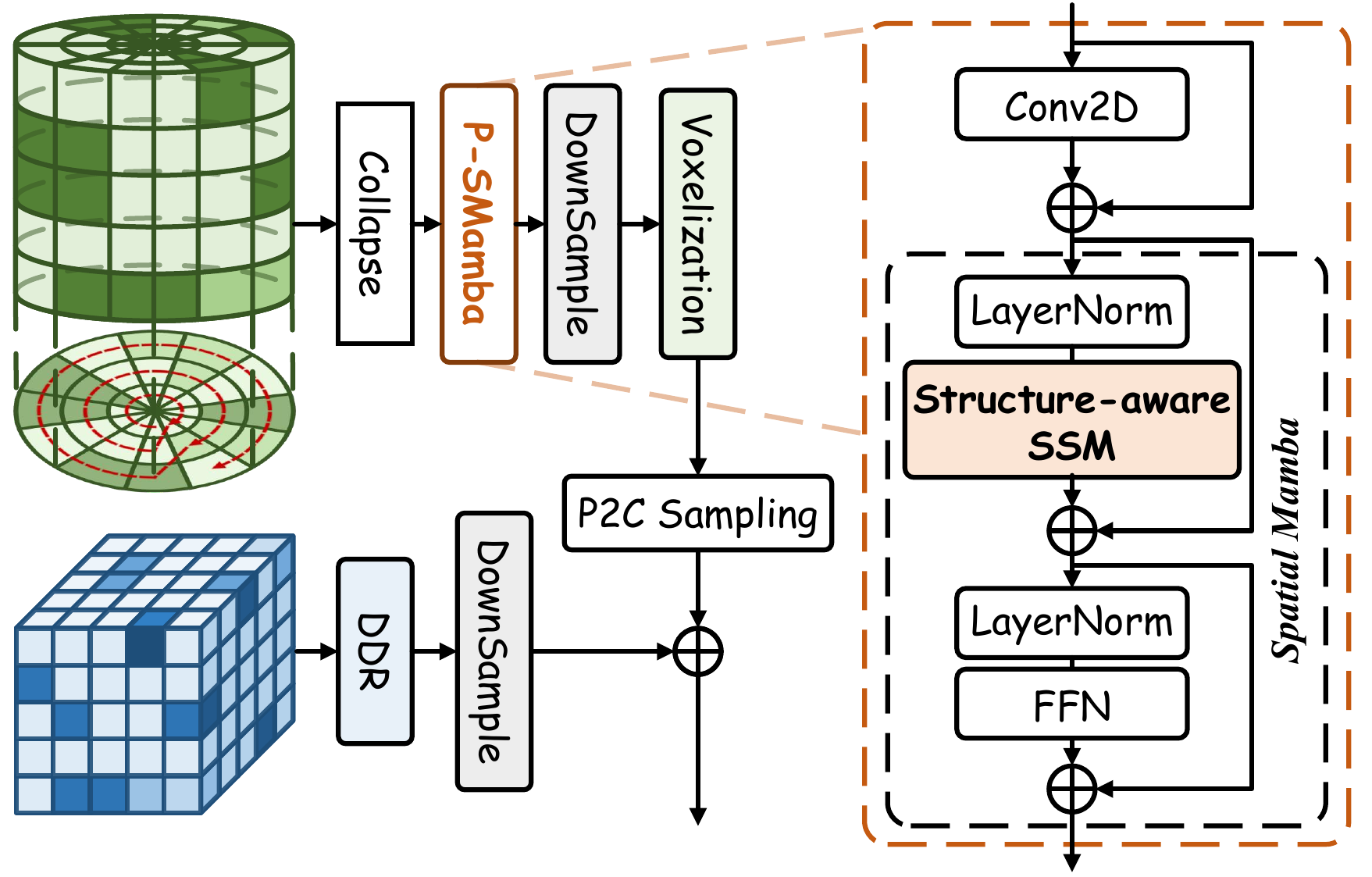}
    \caption{
    \textbf{Polar-spiral Mamba} module utilizes a dual-branch architecture to effectively model the spatial structure of omnidirectional images. 
    Within the cylindrical branch, P-SMamba scans the space in an outward spiral pattern, capturing spatial dependencies associated with the radial variation of visual evidence. 
    Voxel features are progressively aggregated across polar and Cartesian coordinates to generate comprehensive features that maintain geometric and semantic continuity.
    }
    \label{fig:psm}
\end{figure}

Due to the limitations of angular partitioning, cylindrical voxels exhibit inherent discontinuity in data format, leading to a disruption in spatial continuity near the poles, which contradicts the original intent (\ie, spatial perception originates from the center, with higher precision in closer regions). 
Standard 3D convolution struggles to adapt to this critical characteristic, and the high computational cost of transformers makes them difficult to deploy on resource-constrained embodied platforms.
To preserve the continuity of angular space effectively, as shown in Fig.~\ref{fig:psm}, we propose the Polar-spiral Mamba (PsM) module with a dual-branch framework to better utilize the intrinsic spatial structure of omnidirectional images.

We build P-SMamba upon Spatial-Mamba~\citep{xiao2024spatial}, which offers transformer-like long sequence modeling capabilities with linear complexity, to capture the intricate structures within the polar coordinate system.
Specifically, the cylindrical voxel representation $\mathbf{V}_{p} \in \mathbb{R}^{C\times R\times P\times Z}$ is compressed along the vertical dimension into Bird's-Eye View (BEV) feature $\mathbf{B}_{p} \in \mathbb{R}^{C\times R\times P}$, reducing spatial redundancy and improving feature representation efficiency. P-SMamba then scans the polar space in a spiral pattern (see the red dashed line in Fig.~\ref{fig:psm}), starting from the pole and progressively increasing the radius. 
The scanning path aligns well with the properties of omnidirectional imaging, especially in the south pole regions, where information density varies progressively. It effectively captures rich spatial geometries and semantics in a progression from nearby to distant regions.

In each layer, the polar BEV feature $\mathbf{B}_{p}^{i} \in \mathbb{R}^{C_{i}\times R_{i}\times P_{i}}$ at the $i$-th stage is voxelized into $\mathbf{V}_{p}^{i} \in \mathbb{R}^{C_{i}\times R_{i}\times P_{i}\times Z_{i}}$ through reshaping operation; next, using the pre-computed Cartesian-to-cylindrical projection relationship, we resample the cylindrical voxel centroids ($c$) into the corresponding cubic space. Afterwards, the fused voxel feature $\mathbf{V}_{f}^{i}$ is derived by aggregating cube voxel representation $\mathbf{V}_{c}^{i} \in \mathbb{R}^{C_{i}\times H_{i}\times W_{i}\times D_{i}}$ and the gathered voxel feature. The detailed procedure at the $i$-th stage is formulated as follows:
\begin{equation}
    \mathbf{V}_{f}^{i} = \mathbf{V}_{c}^{i} + \Phi_{\rho(c)}(\mathbf{V}_{p}^{i}),~~\text{if $i$} > 1,
\end{equation}
where $\Phi_{a}(b)$ is the sampling of $b$ at Cartesian coordinates $a$, and $\rho(\cdot)$ is the coordinate projection. 
By exploiting the complementary properties of cylindrical and Cartesian voxel representations, PsM effectively models omnidirectional spatial dependencies while retaining the regular metric structure of occupancy space.

\subsection{Occupancy Cost Aggregation}
\label{sec:occupancy_cost_aggregation}

\begin{figure*}[pos=!t]
    \centering
    \includegraphics[width=.95\textwidth]{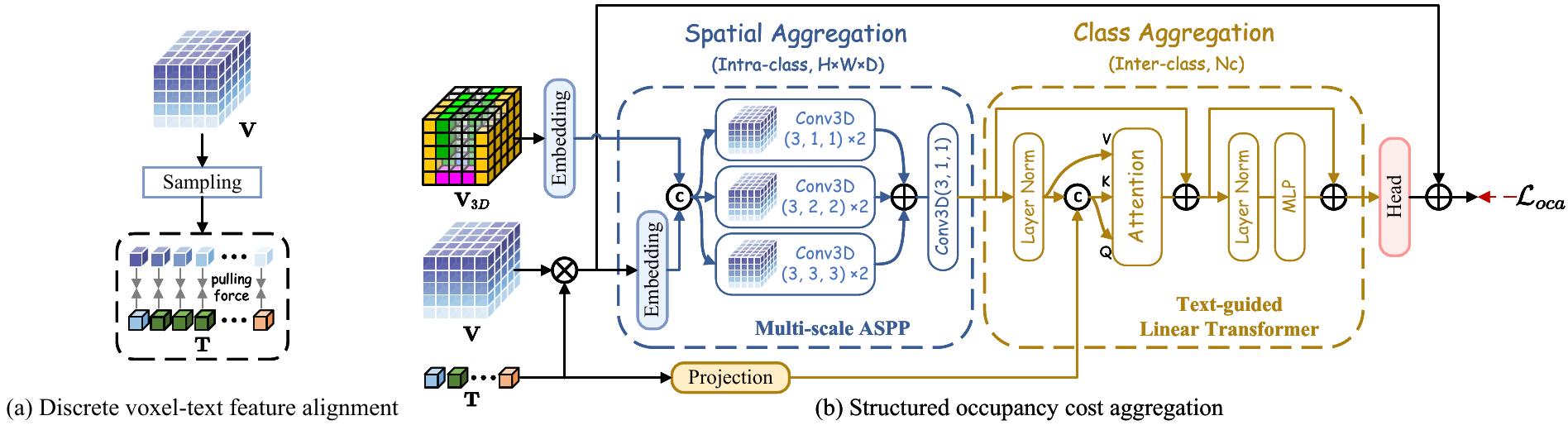}
    \caption{
    \textbf{Illustration of discrete voxel-text feature alignment and the proposed occupancy cost aggregation.}
    (a) Existing discrete voxel-text alignment samples voxel embeddings from $\mathbf{V}$ and directly pulls them closer to their corresponding text embeddings in the shared embedding space, providing point-wise semantic alignment but lacking structured spatial and class-level reasoning. 
    (b) Given occupancy voxel embeddings $\mathbf{V}$, 3D voxel features $\mathbf{V}_{3D}$, and text embeddings $\mathbf{T}$, OCA constructs semantic occupancy cost representations and refines them through two complementary stages: intra-class spatial aggregation over the $H \times W \times D$ voxel space, and inter-class aggregation over $N_c$ semantic categories. The resulting aggregated costs are supervised by $\mathcal{L}_{\mathrm{oca}}$ to enhance geometry-aware voxel-text alignment and open-vocabulary occupancy prediction.
    }
    \label{fig:aggregator}
\end{figure*}

The non-uniform distribution of visual evidence and imbalanced voxel sampling can bias direct feature alignment toward observed base categories, resulting in misalignment of the joint embedding space for unseen classes. 
To mitigate this issue, we propose constructing a refined occupancy cost to characterize the relationship between voxel and text embeddings, thereby avoiding direct discrete feature alignment.

Analogous to the image–text matching cost~\citep{cho2024cat_seg}, the occupancy cost is defined as the similarity between 3D voxel embeddings $\mathbf{V}$ and text embeddings $\mathbf{T}$, typically measured via cosine similarity:
\begin{equation}
    C(s,l) = \frac{V_{s}\cdot T_{l}}{\vert\vert V_{s}\vert\vert~\vert\vert T_{l}\vert\vert},
\end{equation}
where $s$ and $l$ denote 3D spatial positions of the voxel embedding and an index for a class, respectively. 
Intuitively, the resulting occupancy cost volume encodes a set of coarse class-specific 3D semantic response maps, reflecting the spatial distribution of voxel-text semantic affinities. 
Then, we feed the occupancy cost to a single 3D convolution layer, which processes each cost slice $C(:,l)\in\mathbb{R}^{1 \times H\times W\times D}$ independently, to obtain initial occupancy cost embedding $\mathbf{O}_{c}\in\mathbb{R}^{N_{c} \times d_{o} \times H\times W\times D}$. $N_{c}$ represents the number of cost slices, and $d_{o}$ is the embedding dimension.

To promote spatial coherence and capture multi-scale local context, as illustrated in Fig.~\ref{fig:aggregator}, we apply Atrous Spatial Pyramid Pooling (ASPP)~\citep{chen2018deeplabv3+} independently to the cost embeddings of each semantic class.
The parallel atrous convolutions aggregate spatial evidence over multiple receptive fields while preserving class-specific cost representations.
Subsequent to spatial aggregation, a linear transformer~\citep{katharopoulos2020transformers} block for class aggregation is applied to consider the text modality, explicitly capturing relationships among different classes. We further leverage the voxel feature $\mathbf{V}_{3D}$ and text embeddings $\textbf{T}$ to provide spatial structure or contextual information of the input, thereby guiding the overall process. Ultimately, the output is fed into the prediction head, with a residual connection for final inference. 

However, voxel-wise cross-entropy supervises each spatial location independently and does not explicitly optimize scene-level class consistency. In our experiments, directly applying cross-entropy loss to the aggregated occupancy costs tends to produce spatially fragmented class responses, which limit generalization to novel categories. We therefore utilize the scene-class affinity loss~\citep{cao2022monoscene} to jointly optimize the class-wise derivable precision, recall, and specificity of the cost predictions over valid voxels.
Considering the predicted probabilities of occupancy cost aggregation $p$ and the ground truth class of voxel embeddings $y$, the loss $\mathcal{L}_{\mathrm{oca}}$ is defined as follows: 
\begin{equation}
    \mathcal{L}_{\mathrm{oca}} = -\frac{1}{N_{c}} \sum\limits_{l=1}^{N_{c}}(P_{l}(p,y)+R_{l}(p,y)+S_{l}(p,y)).
\end{equation}
$P_{l}$ (Precision) and $R_{l}$ (Recall) are the performance measures of similar class $l$ voxels, whereas $S_{l}$ (Specificity) measures the performance of dissimilar voxels (\ie, not of class $l$).
In practice, during training, the $\mathcal{L}_{\mathrm{oca}}$ is computed only over valid voxels belonging to the base classes.
The resulting scene-level supervision encourages more coherent class responses in the aggregated occupancy costs and improves open-vocabulary occupancy prediction.

\subsection{Natural Modality Alignment}
\label{sec:natural_modality_alignment}

Despite its strong generalization capabilities, CLIP struggles with the inherent domain gap between image and text embeddings, even after extensive alignment during training~\citep{zhao2025dpseg,tan2023ovo}. 
This discrepancy is further exacerbated by the projection errors inherent to panoramic images. Under the conventional pixel-voxel feature alignment, the \textit{``pixel-voxel-text''} semantic consistency cannot be effectively guaranteed, thereby struggling to recognize novel semantics. 
Moreover, since only base class semantics are available during training, directly employing a learning-based alignment strategy may inevitably develop excessive reliance on the distribution of seen semantics while deteriorating the ability to understand unlimited semantics. 

To overcome these issues, we propose a gradient-free natural modality gap alignment strategy between text embeddings and semantic prototypes prior to occupancy cost aggregation. Specifically, it iteratively aggregates text embeddings with prototypes of seen semantics, which are obtained by filtering the base class pixel embeddings with the Exponential Moving Average (EMA) method~\citep{cai2021exponential} during the training phase, achieving co-optimization in shared embedding space until convergence.
During training, the corresponding text embeddings and initialized prototypes can be represented by $\mathbf{T}=\{T_{l}\in\mathbb{R}^{d}\}^{L_{b}}_{l}$ and $\mathbf{P}=\{P_{l}\in\mathbb{R}^{d}\}^{L}_{l}$, $L=L_{b}+K_{n}$. It is worth noting that we also introduce class-agnostic learnable auxiliary prototypes and initialize them to implicitly capture the unseen semantics. Here, $K_n$ denotes the number of learnable auxiliary prototypes. 
For convenience, we omit the class index $l$ in the example. 
Firstly, the base prototypes $\mathbf{P}^{b}$ are updated by leveraging the image feature $\mathbf{f}_{seg}$ through the EMA method:
\begin{equation}
    \mathbf{P}^{b}_{t} = \alpha\cdot\mathbf{P}^{b}_{t-1} + (1-\alpha)\cdot(\frac{1}{|\Omega_{b}|}\sum\limits_{i\in\Omega_{b}}\mathbf{f}_{seg}(i)),
\end{equation}
where $t$ denotes the current training iteration, $\alpha\in[0,1]$ is the update rate and $\Omega_{b}$ is the valid pixel embedding set. 

Then, we define the affinity $\mathcal{S}$ by calculating cosine similarity between $\mathbf{T}_{t}^{0}$ and $\mathbf{P}_{t}^{0}$ at the $0$-th step of the Random Walk~\citep{grady2006random}:
\begin{equation}
  \mathcal{S} = \lambda\frac{\mathbf{T}_{t}^{0}\cdot\mathbf{P}_{t}^{0}}{\vert\vert \mathbf{T}_{t}^{0}\vert\vert~\vert\vert \mathbf{P}_{t}^{0}\vert\vert},
  \label{eq:sim}
\end{equation}
where $\lambda$ is the scaling factor. At the $k$-th iteration step, the prototypes $\mathbf{P}_{t}^{k}$ are derived by the original prototypes $\mathbf{P}_{t}^{0}$ and the updated text embeddings $\mathbf{T}_{t}^{k-1}$ from previous iteration:
\begin{equation}
  \mathbf{P}_{t}^{k} = \beta(\mathcal{S})^{\top}\mathbf{T}_{t}^{k-1} + (1-\beta)\mathbf{P}_{t}^{0}.
\end{equation}
Subsequently, it is utilized to update the text embeddings of this step:
\begin{equation}
  \mathbf{T}_{t}^{k} = \beta\mathcal{S}~\mathbf{P}_{t}^{k} + (1-\beta)\mathbf{T}_{t}^{0},
\end{equation}
where $\beta\in(0,1)$ is the degree that controls the domain alignment from the current state. By combining the above steps, we can obtain the overall walk formula of $\mathbf{T}_{t}^{k}$ in a unified and compact expression:
\begin{equation}
  \mathbf{T}_{t}^{k} = (\beta^{2}\mathcal{A})^{k}~\mathbf{T}_{t}^{0} + (1-\beta)\sum\limits_ {j=0}^{k-1}(\beta^{2}\mathcal{A})^{j}(\beta\mathcal{S}~\mathbf{P}_{t}^{0} + \mathbf{T}_{t}^{0}),
\end{equation}
\begin{equation}
  \mathcal{A} = \mathcal{S}(\mathcal{S})^{\top},
  ~\beta^2\|\mathcal{S}\|_2^2<1.
\end{equation}
Mathematically, the whole walk process can be presented in closed form based on the Neumann Series~\citep{meyer2023matrix} when $k \to \infty$:
\begin{equation}
  \mathbf{T}_{t}^{\infty} = (1-\beta)(\mathbf{I}-\beta^{2}\mathcal{A})^{-1}(\beta\mathcal{S}~\mathbf{P}_{t}^{0} + \mathbf{T}_{t}^{0}),
\end{equation}
where $\mathbf{I}$ is the identity matrix. Finally, $\mathbf{T}_{t}^{\infty}$ serves as the optimized text embeddings for the subsequent cost construction.

\begin{figure*}[!t]
    \centering
    \includegraphics[width=.8\textwidth]{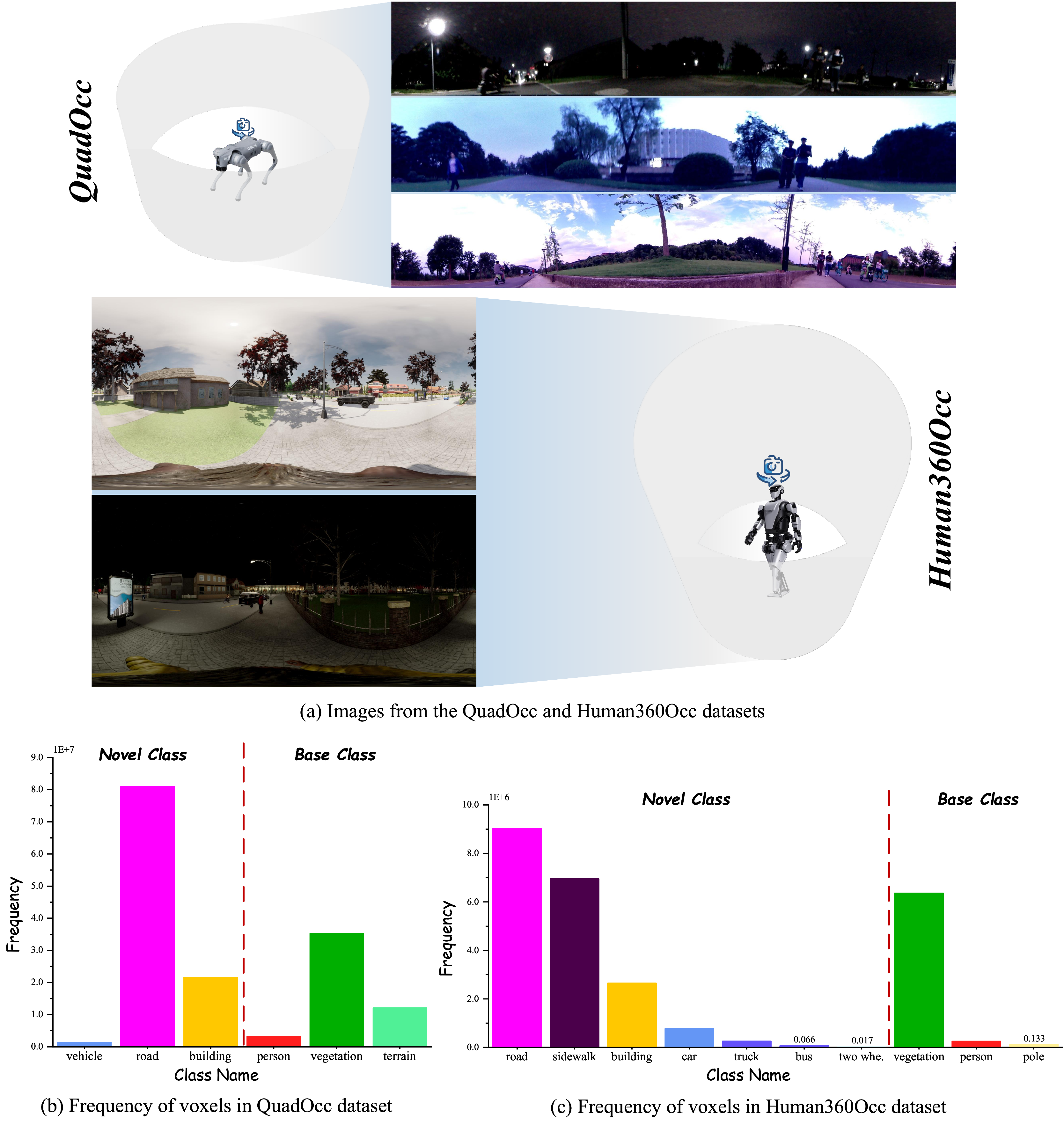}
    \caption{
    \textbf{Images and the frequency of voxels for each semantic category} in the QuadOcc~\citep{shi2026oneocc} and Human360Occ~\citep{shi2026oneocc} datasets.
    }
    \label{fig:datasets}
\end{figure*}

\subsection{Training and Inference}
\subsubsection{Training} 
In this work, MonoScene~\citep{cao2022monoscene} is chosen as the primary semantic occupancy network due to its efficiency and flexibility~\citep{shi2026oneocc}. 
Notably, to ensure transferability, O3N is designed to be a modular and general framework built atop a variety of structurally similar fully supervised semantic occupancy architectures like MonoScene~\citep{cao2022monoscene} and SGN~\citep{mei2024sgn}. 
For voxels belonging to novel classes, we solely possess their geometric positions without semantic information. 
Consequently, we aggregate all novel classes into a new class, uniformly referred to as the unknown class. 
Using the base class labels during training corresponds to incorporating $L_{b} +1$ semantic labels, referred to as processed semantic labels. 
During training, we employ $\mathcal{L}_{occ}$~\citep{cao2022monoscene} to supervise the semantic occupancy head:
\begin{equation}
    \mathcal{L}_{occ} = \mathcal{L}_{ce} + \mathcal{L}^{sem}_{scal} + \mathcal{L}^{geo}_{scal} + \mathcal{L}_{fp}.
\end{equation}
For the open-vocabulary learning, we utilize $\mathcal{L}_{vox-pix}$ from OVO~\citep{tan2023ovo} and $\mathcal{L}_{oca}$ to align voxel, pixel, and textual features. Therefore, the total loss is:
\begin{equation}
    \mathcal{L}=\mathcal{L}_{occ} + \mathcal{L}_{vox-pix} + \mathcal{L}_{oca}.
\end{equation}

\subsubsection{Inference} 
For the base classes, we directly use the occupancy head for reliable semantic occupancy prediction. For the novel classes, we derive $\mathbf{V}$ via the distillation module and generate the novel class logit by combining the similarity scores between $\mathbf{V}$ and the $\{T_{l}\}^{L_{n}}_{l}$ with the predicted logits $p$ from the OCA for accurate occupancy prediction.

\section{Experiments}
\label{sec:experiments}

\subsection{Experimental Setups}
\subsubsection{Datasets} 
We evaluate methods on two challenging omnidirectional occupancy datasets: QuadOcc~\citep{shi2026oneocc} and Human360Occ~\citep{shi2026oneocc}. Fig.~\ref{fig:datasets} illustrates sample omnidirectional images from the QuadOcc and Human360Occ datasets, along with the corresponding voxel frequency distributions across semantic classes, providing a clear overview of their scene diversity and inherent class imbalance.
QuadOcc is a real-world first-person $360^{\circ}$ occupancy dataset on a quadruped robot within a campus environment.
It contains $7$ scenes for training, $3$ scenes for validation/test, totaling $24K$ frames with standardized day/dusk/night coverage. It has $6$ semantic classes with an additional \emph{empty} class on a $64{\times}64{\times}8$ grid ($0.4m$). Each sample covers a range of $[{-}12.8m,{-}12.8m,{-}1.2m,12.8m,12.8m,2.0m]$, with a fine voxel size of $[0.4m,0.4m,0.4m]$ on a $64{\times}64{\times}8$ grid, reflecting embodied compute and lower speed than cars.
Human360Occ (H3O) is a CARLA-based human-ego $360^{\circ}$ occupancy dataset with simulated gait. H3O contains $160$ sequences, totaling $8K$ frames over $16$ maps and multiple weathers/times. Each sample covers a range of $[{-}12.8m,{-}12.8m,{-}2.4m,12.8m,12.8m,0.8m]$, and provides semantic occupancy labels at native resolutions of $64{\times}64{\times}8$. It uses $10$ classes along with an \emph{empty} class for training, supports \emph{within-city} (Homo, per-map $8{:}2$ train/val) and \emph{cross-city} (Heter, default $12{/}4$ maps) splits.

\subsubsection{Class Split} 
Following prior open-vocabulary occupancy protocols~\citep{tan2023ovo,li2025ago}, we partition the semantic categories of each dataset into base and novel sets. 
For QuadOcc, \emph{vehicle}, \emph{road}, and \emph{building} are designated as novel classes, together accounting for approximately $68\%$ of the annotated non-empty voxels. The remaining categories, namely \emph{person}, \emph{vegetation}, and \emph{terrain}, are treated as base classes.
For Human360Occ, we designate \emph{road}, \emph{sidewalk}, \emph{building}, \emph{car}, \emph{truck}, \emph{bus}, and \emph{two-wheeler} as novel classes, collectively accounting for approximately $75\%$ of the annotated non-empty voxels. The remaining categories, including \emph{vegetation}, \emph{person}, and \emph{pole}, constitute the base-class set.
Notably, the base-class distributions are highly imbalanced, with several categories representing less than $1\%$ of the non-empty voxels. This long-tailed supervision makes it challenging to learn robust base-class representations and to calibrate predictions across base and novel categories.
Nevertheless, restricting category-specific supervision to a limited base vocabulary provides a label-efficient setting that is amenable to future data and task expansion.

\begin{table*}[pos=!t]
    \centering
    \caption{\textbf{Omnidirectional 3D occupancy prediction results on the QuadOcc~\citep{shi2026oneocc} validation set.} \textbf{Bold} $=$ best, \underline{underline} $=$ second best (within the Fully Supervised block; Open-vocabulary block marked separately). L: LiDAR; C: Camera. 
    }
    \label{tab:comparison_quadssc}
    \resizebox{.7\textwidth}{!}{
    \begin{tabular}{l|c|c|cccc|cccc}
        \toprule
        & & & \multicolumn{4}{c}{\textbf{Novel Class}}& \multicolumn{4}{c}{\textbf{Base Class}}\\
        \textbf{Method}& \textbf{Input}& \textbf{mIoU}& \rotatebox{90}{\textcolor{colorVehicle}{$\blacksquare$} \makecell[l]{vehicle \\ (\quadoccfreq{vehicle}\%)}}
        & \rotatebox{90}{\textcolor{colorRoad}{$\blacksquare$} \makecell[l]{road \\ (\quadoccfreq{road}\%)}}
        & \rotatebox{90}{\textcolor{colorBuilding}{$\blacksquare$} \makecell[l]{building \\ (\quadoccfreq{building}\%)}}
        & mean 
        & \rotatebox{90}{\textcolor{colorPedestrian}{$\blacksquare$} \makecell[l]{person \\ (\quadoccfreq{pedestrian}\%)}} 
        & \rotatebox{90}{\textcolor{colorVegetation}{$\blacksquare$} \makecell[l]{vegetation \\ (\quadoccfreq{vegetation}\%)}}
        & \rotatebox{90}{\textcolor{colorTerrain}{$\blacksquare$} \makecell[l]{terrain \\ (\quadoccfreq{terrain}\%)}}
        & mean \\
        \midrule
        \hline
        \rowcolor{mygray} \multicolumn{11}{c}{\emph{\textbf{Fully-supervised}}}\\
        \midrule
        SSCNet~\citep{song2017semantic}& \emph{L}& 14.60& 0.00& 44.34& \underline{16.06}& 20.13& 0.04& 22.41& 4.77&9.07\\
        SSCNet-full~\citep{song2017semantic}& \emph{L}& 16.38& 0.71& 55.14& 14.93& 23.59& 0.04& 19.66& 7.76&9.15\\
        LMSCNet~\citep{roldao2020lmscnet}& \emph{L}& 18.44& 0.88& \textbf{57.02}& \textbf{16.45}& 24.78& 0.20& 24.39& \underline{11.70}&12.10\\
        OccRWKV~\citep{wang2025occrwkv}& \emph{L}& 16.69& 2.05& 54.17& 10.29& 22.17& 0.35& \underline{24.92}& 8.38&11.22\\
        VoxFormer-S~\citep{li2023voxformer}& \emph{C}& 9.36& 0.31& 36.83& 6.48& 14.54& 1.67& 7.09& 3.81&4.19\\
        OccFormer~\citep{zhang2023occformer}& \emph{C}& 13.02& 0.29& 49.46& 10.36& 20.04& 0.37& 15.00& 2.64&6.00\\
        SGN-S~\citep{mei2024sgn}& \emph{C}& 17.17& 7.39& 53.35& 9.06& 23.27& 1.88& 21.95& 9.39&11.07\\
        SGN-T~\citep{mei2024sgn}& \emph{C}& 17.50& \underline{8.23}& 52.74& 10.47& 23.81& \underline{2.09}& 23.04& 8.45&11.19\\
        MonoScene~\citep{cao2022monoscene}& \emph{C}& \underline{19.19}& 8.15& \underline{55.66}& 12.88& \underline{25.56}& 1.59& \textbf{26.10}& 10.78& \underline{12.82}\\
        OneOcc~\citep{shi2026oneocc} & \emph{C}& \textbf{20.56}& \textbf{12.16}& 54.41& 16.03& \textbf{27.53}& \textbf{2.86}& 24.91& \textbf{13.01}& \textbf{13.59}\\
        \hline
        \rowcolor{mygray} \multicolumn{11}{c}{\emph{\textbf{Open-vocabulary}}}\\
        \midrule
        OVO (SGN-S)~\citep{tan2023ovo}& \emph{C}& 13.81& \textbf{2.38}& 44.78& 5.64& 17.60& 0.24& 19.64& 10.17& 10.02\\
        {\cellcolor{mygreen!10}}\textbf{Ours} (SGN-S)& 
        {\cellcolor{mygreen!10}}\emph{C}&
        {\cellcolor{mygreen!10}}\underline{15.52}&
        {\cellcolor{mygreen!10}}0.56&
        {\cellcolor{mygreen!10}}\underline{52.08}&
        {\cellcolor{mygreen!10}}\underline{7.88}&
        {\cellcolor{mygreen!10}}\underline{20.18}&
        {\cellcolor{mygreen!10}}\textbf{2.28}&
        {\cellcolor{mygreen!10}}\textbf{21.29}&
        {\cellcolor{mygreen!10}}9.03&
        {\cellcolor{mygreen!10}}\underline{10.87}\\
        OVO~\citep{tan2023ovo}& \emph{C}& 14.33& \underline{2.11}& 46.34& 6.00& 18.15& 0.39& 18.84& \underline{12.32}& 10.52\\
        {\cellcolor{mygreen!10}}\textbf{Ours}& 
        {\cellcolor{mygreen!10}}\emph{C}&
        {\cellcolor{mygreen!10}}\textbf{16.54}&
        {\cellcolor{mygreen!10}}0.52&
        {\cellcolor{mygreen!10}}\textbf{54.22}&
        {\cellcolor{mygreen!10}}\textbf{8.76}&
        {\cellcolor{mygreen!10}}\textbf{21.16}&
        {\cellcolor{mygreen!10}}\underline{0.79}&
        {\cellcolor{mygreen!10}}\underline{20.17}&
        {\cellcolor{mygreen!10}}\textbf{14.80}&
        {\cellcolor{mygreen!10}}\textbf{11.92}\\
        \bottomrule
    \end{tabular}
    }
\end{table*}

\subsubsection{Evaluation Metric} 
As we aim at omnidirectional open-vocabulary occupancy prediction, our primary performance metric is the mean Intersection over Union (mIoU). 
For the benchmark, we report the mIoU across all classes, as well as separately for novel and base classes.

\subsubsection{Implementation Details} 
Our proposed method is not tied to any specific architecture and can be readily adapted to various occupancy prediction networks that incorporate visual inputs.
We assess its generality by applying it to two representative models: MonoScene~\citep{cao2022monoscene} and SGN~\citep{mei2024sgn}, keeping the parameters consistent with each original model. Among them, MonoScene~\citep{cao2022monoscene} achieves the best performance and is therefore selected as the primary body of our method. To ensure a fair comparison, we adopt the parameter configuration of MonoScene~\citep{cao2022monoscene}, in line with OVO~\citep{tan2023ovo}, and adjust the weight decay coefficient from $1{\times}10^{-4}$ to $1{\times}10^{-3}$ to mitigate overfitting. The model is trained for $25$ epochs on $4$ NVIDIA RTX 3090 GPUs with a total batch size of $4$. 
In the PsM module, the cylindrical voxels with a shape of ($R{:}32$, $P{:}90$, $Z{:}8$), representing ($r$, $\theta$, $z$) in polar coordinates, and the depths of P-SMamba are set to $2$ and $4$. 
The distillation module consists of $4$ convolution layers, which replace the MLPs in OVO and are used to align voxel features to $512$ dimensions. 
We utilize the text encoder from CLIP~\citep{radford2021learning} to extract textual features, with an embedding dimension of $512$; in addition, following LSeg~\citep{li2022language}, we extract high-resolution pixel-level image features $\mathbf{f}_{seg}$. 
The hyperparameters $d_{o}$, $\alpha$, $\lambda$, and $\beta$ are set to $128$, $0.9$, $1.0$, and $0.1$, respectively.

\begin{table*}[pos=!t]
    \centering
    \caption{\textbf{Omnidirectional 3D occupancy prediction results on H3O~\citep{shi2026oneocc} Homo (within-city) split.} C: Camera.}
    \label{tab:comparison_h3o}
    \resizebox{\textwidth}{!}{
    \begin{tabular}{l|c|c|cccccccc|cccc}
        \toprule
        & & & \multicolumn{8}{c}{\textbf{Novel Class}}& \multicolumn{4}{c}{\textbf{Base Class}}\\
        \textbf{Method}& \textbf{Input}& \textbf{mIoU}
        & \ColHeadHomo{hthreeoRoad}{road}{road} 
        & \ColHeadHomo{hthreeoSidewalk}{sidewalk}{sidewalk} 
        & \ColHeadHomo{hthreeoBuilding}{building}{building}
        & \ColHeadHomo{hthreeoCar}{car}{car} 
        & \ColHeadHomo{hthreeoTruck}{truck}{truck}
        & \ColHeadHomo{hthreeoBus}{bus}{bus} 
        & \ColHeadHomo{hthreeoTwoWheeler}{two\_wheeler}{twoWheeler} 
        & mean 
        & \ColHeadHomo{hthreeoVegetation}{vegetation}{vegetation} 
        & \ColHeadHomo{hthreeoPerson}{person}{person} 
        & \ColHeadHomo{hthreeoPole}{pole}{pole} 
        & mean \\
        \midrule
        \hline
        \rowcolor{mygray} \multicolumn{15}{c}{\emph{\textbf{Fully-supervised}}}\\
        \midrule
        VoxFormer-S~\citep{li2023voxformer}& \emph{C}& 11.09& 22.27& 17.44& 2.86& 0.87& 0.14& 0.00& 0.00& 6.22& 5.65& 61.46& 0.18& 22.43\\
        OccFormer~\citep{zhang2023occformer}& \emph{C}& 24.88& 54.11& 46.41& 15.14& 17.57& 4.30& 0.00& 1.65& 19.88& 32.84& 65.08& 11.72& 36.55\\
        SGN-S~\citep{mei2024sgn}& \emph{C}& 28.56& 54.81& 47.44& 15.64& 24.23& 8.53& 2.62& 13.51& 23.83& 33.22& 65.88& 19.72& 39.61\\
        SGN-T~\citep{mei2024sgn}& \emph{C}& 28.64& 54.55& 44.61& 15.51& 23.91& 7.68& 4.52& 15.77& 23.79& 35.16& 65.24& 19.46& 39.95\\
        MonoScene~\citep{cao2022monoscene}& \emph{C}& 33.46& 62.45& 57.18& 19.97& 29.76& 13.77& 5.03& 16.76& 29.27& 38.91& 70.15& 20.67& 43.24\\
        OneOcc~\citep{shi2026oneocc}& \emph{C}& 37.29& 63.82& 62.92& 22.86& 33.74& 17.79& 6.47& 26.58& 33.45& 41.03& 71.13& 26.55& 46.24\\
        \hline
        \rowcolor{mygray} \multicolumn{15}{c}{\emph{\textbf{Open-vocabulary}}}\\
        \midrule
        OVO (SGN-S)~\citep{tan2023ovo}& \emph{C}& 21.40& 39.23& 24.28& 12.49& 16.73& 1.70& 0.62& 0.00& 13.58& 33.90& 66.84& 18.16& 39.63\\
        {\cellcolor{mygreen!10}}\textbf{Ours} (SGN-S)& 
        {\cellcolor{mygreen!10}}\emph{C}&
        {\cellcolor{mygreen!10}}22.45&
        {\cellcolor{mygreen!10}}39.07&
        {\cellcolor{mygreen!10}}26.68&
        {\cellcolor{mygreen!10}}14.51&
        {\cellcolor{mygreen!10}}18.60&
        {\cellcolor{mygreen!10}}0.00&
        {\cellcolor{mygreen!10}}0.00&
        {\cellcolor{mygreen!10}}0.00&
        {\cellcolor{mygreen!10}}14.12&
        {\cellcolor{mygreen!10}}37.15&
        {\cellcolor{mygreen!10}}65.66&
        {\cellcolor{mygreen!10}}22.82&
        {\cellcolor{mygreen!10}}41.88\\
        OVO~\citep{tan2023ovo} & \emph{C}& 23.39& 38.69& 14.30& 18.46& 21.48& 3.67& 0.15& 0.00& 13.82& 40.36& 70.67& 26.17& 45.73\\
        {\cellcolor{mygreen!10}}\textbf{Ours} & 
        {\cellcolor{mygreen!10}}\emph{C}&
        {\cellcolor{mygreen!10}}24.25&
        {\cellcolor{mygreen!10}}42.08&
        {\cellcolor{mygreen!10}}25.55&
        {\cellcolor{mygreen!10}}17.40&
        {\cellcolor{mygreen!10}}22.49&
        {\cellcolor{mygreen!10}}0.00&
        {\cellcolor{mygreen!10}}0.00&
        {\cellcolor{mygreen!10}}0.03&
        {\cellcolor{mygreen!10}}15.36&
        {\cellcolor{mygreen!10}}39.16&
        {\cellcolor{mygreen!10}}70.20&
        {\cellcolor{mygreen!10}}25.61&
        {\cellcolor{mygreen!10}}44.99\\
        \bottomrule
    \end{tabular}
    }
\end{table*}

\begin{figure*}[pos=!t]
    \centering
    \includegraphics[width=\textwidth]{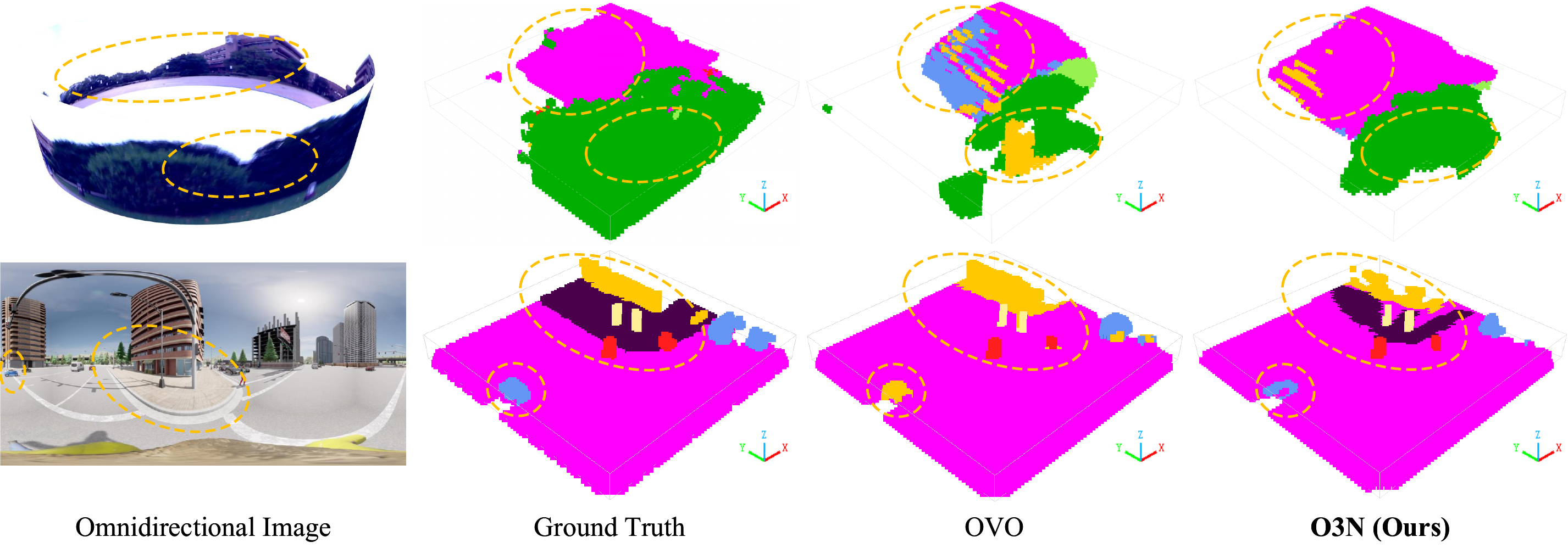}
    \caption{
    \textbf{Qualitative results.} O3N more effectively maintains the clarity and continuity of global geometry and semantics, and achieves significant improvements over the baseline in terms of perception and generalization to unknown semantics.
    }
    \label{fig:visResults}
\end{figure*}

\subsection{Comparison to State-of-the-Art Methods}
On the real-world QuadOcc validation set, our method achieves improvements on both known and novel classes. As shown in Tab.~\ref{tab:comparison_quadssc}, with category-specific semantic supervision restricted to approximately $32\%$ of the annotated non-empty voxels, our approach reaches an overall mIoU of $16.54$, surpassing the OVO~\citep{tan2023ovo} ($14.33$). Notably, the \emph{novel} mIoU rises to $21.16$, outperforming several fully supervised methods (\ie, SSCNet~\citep{song2017semantic}, OccFormer~\citep{zhang2023occformer}, and VoxFormer-S~\citep{li2023voxformer}). These results demonstrate that, even under limited supervision, our framework can effectively extract and transfer rich panoramic semantic structures. Moreover, when adapted to the SGN-S~\citep{mei2024sgn} backbone, our framework continues to deliver consistent performance gains. The performance gains from $13.81$ to $15.52$ mIoU, with both novel and base-class averages showing notable enhancement. 
Collectively, these results highlight the robustness and transferability of the proposed framework, supporting its potential for open-world perception in intelligent systems. We further evaluate our framework on the H3O dataset, which simulates human-ego panoramic perception. As reported in Tab.~\ref{tab:comparison_h3o}, our method consistently outperforms all open-vocabulary counterparts and achieves results comparable to several fully supervised methods. Under the open-vocabulary setting, our framework attains an overall mIoU of $24.25$. 

\begin{figure*}[pos=!t]
    \centering
    \includegraphics[width=\textwidth]{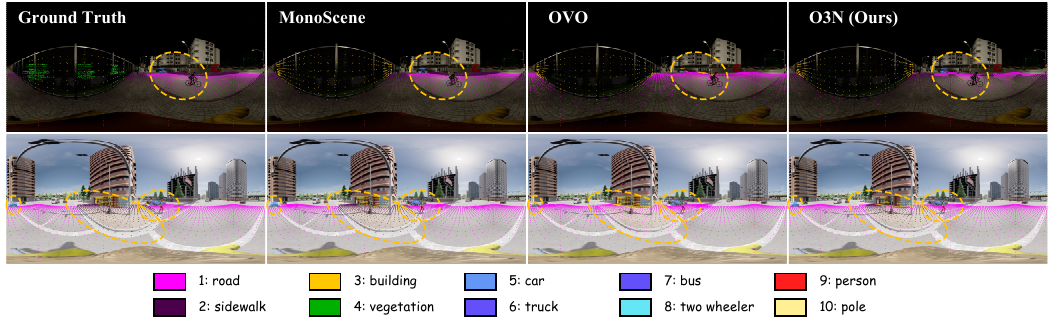}
    \caption{
    \textbf{Qualitative comparison of semantic occupancy projections on H3O~\citep{shi2026oneocc}.} We project semantic voxel centers from the ground truth and different methods onto the corresponding equirectangular omnidirectional images. Colored points indicate semantic categories, as shown in the legend. Yellow dashed circles highlight challenging regions involving local object structures, road/sidewalk boundaries, and small dynamic objects. Compared with MonoScene~\citep{cao2022monoscene} and OVO~\citep{tan2023ovo}, O3N produces semantic occupancy projections that are more consistent with the ground truth, with cleaner traversable-surface layouts, more coherent structural boundaries, and better preservation of object-level details.
    }
    \label{fig:projVis}
\end{figure*}

The qualitative results in Fig.~\ref{fig:visResults} further demonstrate that our approach effectively captures omnidirectional spatial geometry and maintains semantic continuity. 
Moreover, to examine the qualitative behavior of different methods, Fig.~\ref{fig:projVis} presents a projected-view comparison on the H3O dataset, where semantic voxel centers are projected onto the original omnidirectional images. This visualization provides an intuitive view of both geometric occupancy quality and semantic consistency in the omnidirectional scene. As shown in the highlighted regions, the baseline often introduces fragmented predictions and semantic leakage around object boundaries. In contrast, O3N yields more coherent road and sidewalk layouts, sharper building structures, and better localized predictions for small foreground objects. These observations indicate that O3N not only improves the overall mIoU, but also better preserves spatial continuity and object-level semantic details in panoramic 3D scene understanding.

Interestingly, the relatively higher IoU achieved by baselines on certain rare categories does not necessarily indicate more reliable semantic recognition. As illustrated in Fig.~\ref{fig:visResults}, OVO tends to produce spatially overextended erroneous semantic predictions, which increase the overlap with ground-truth voxels of rare categories despite introducing substantial false positives. In contrast, O3N promotes spatially coherent semantic predictions and suppresses spurious responses, potentially at the expense of reduced recall for rare categories with insufficient visual evidence. These observations highlight the remaining challenge of balancing geometric-semantic consistency with rare-category recall under highly imbalanced semantic distributions.

\subsection{Ablation Studies}
To verify the effectiveness of O3N, we conduct comprehensive ablation studies based on the QuadOcc dataset to dissect core module contributions.

\subsubsection{Main Components} 
As summarized in Tab.~\ref{tab:ablation_components}, introducing the PsM alone yields consistent improvements across all three metrics (${+}0.53$ for \emph{Novel} mIoU, ${+}0.15$ for \emph{Base} mIoU, and ${+}0.34$ for overall mIoU). By modeling spatial priors in polar coordinates, PsM effectively preserves geometric continuity and improves voxel representation quality; 
notably, thanks to the linear complexity of Mamba~\citep{xiao2024spatial}, PsM incurs negligible memory overhead in practice. 
Furthermore, since OCA rectifies feature misalignment and strengthens voxel-wise geometric consistency, incorporating OCA further increases \emph{Novel} mIoU and overall mIoU to $19.78$ and $15.40$, achieving additional gains of ${+}1.19$ and ${+}0.58$, respectively. 
With the introduction of NMA, the model achieves the best performance, reaching $21.16$ for \emph{Novel} mIoU and $16.54$ for overall mIoU, demonstrating that the proposed cross-modality alignment mechanism effectively bridges the image–text domain gap and enhances generalization to both seen and unseen classes. 
In terms of efficiency, while OCA and NMA introduce modest computational costs, they yield significant performance improvements that fully justify the overhead. The full model (O3N) achieves $9.41$ FPS with only $4.97$ GB memory during inference, maintaining competitive inference efficiency.

\begin{table}[pos=!t]
    \centering
    \caption{\textbf{Ablation study of each component.} PsM: Polar-spiral Mamba; OCA: Occupancy Cost Aggregation; NMA: Natural Modality Alignment.}
    \resizebox{\linewidth}{!}{
    \begin{tabular}{c|cc|cc|c|cc}
        \toprule
        PsM& OCA& NMA& \emph{Novel} mIoU& \emph{Base} mIoU& mIoU & FPS~$\uparrow$ & Mem. (GB)~$\downarrow$ \\
        \midrule
        & & & 18.06& 10.90& 14.48 & 10.67 & 4.28\\
        \midrule
        \ding{51}& & & 18.59& 11.05& 14.82 & 9.98& 4.31\\
        \ding{51}& \ding{51}& & 19.78& 11.02& 15.40 & 9.71& 4.86\\
        \ding{51}& \ding{51}& \ding{51}& \textbf{21.16}& \textbf{11.92}& \textbf{16.54} & 9.41& 4.97\\
        \bottomrule
    \end{tabular}
    }
    \label{tab:ablation_components}
\end{table}

\subsubsection{Analysis of PsM}
We conduct a component-wise ablation study in Tab.~\ref{tab:ablation_PsM} to examine the contributions of the cylindrical representation, P-SMamba, and the spiral scanning pattern.
Replacing the cylindrical representation with a Cartesian counterpart reduces the mIoU, indicating that the sensor-centered cylindrical parameterization is better suited to modeling the radial and azimuthal structure of omnidirectional scenes.
When P-SMamba is retained but spiral scanning is replaced with the alternative traversal strategy, the performance further decreases, supporting the importance of the scan order for organizing long-range dependencies in the polar space.
The complete configuration, which combines the cylindrical representation, P-SMamba, and spiral scanning, achieves the highest performance among the evaluated variants.
These results suggest that the improvements primarily stem from our geometric and structural designs rather than merely from an increase in parameter count.

\begin{table}[pos=!t]
    \centering
    \caption{\textbf{Ablation study of PsM.}}
    \label{tab:ablation_PsM}
    \resizebox{\linewidth}{!}{%
    \begin{tabular}{@{}ccc|cc|c@{}}
        \toprule
        Cylin. Voxel& P-SMamba& Spiral scan.& \emph{Novel} mIoU& \emph{Base} mIoU& mIoU \\
        \midrule
        \ding{51}& \ding{51}& \ding{51}& \textbf{21.16}& \textbf{11.92}& \textbf{16.54}\\
         & \ding{51}& \ding{51}& 20.04 & 11.32 & 15.68\\
        \ding{51}& \ding{51}& & 19.30 & 11.29 & 15.30\\
        \ding{51}& & & 19.74 & 11.14 & 15.44\\
        \bottomrule
    \end{tabular}
    }
\end{table}

\begin{table}[pos=!t]
    \centering
    \caption{\textbf{Ablation study on cylindrical voxel granularity} under different radial and angular resolutions for the PsM module. 
    Mem. denotes training peak memory.}
    \label{tab:psm_resolution}
    \resizebox{.82\linewidth}{!}{
    \begin{tabular}{cc|cc|cc}
        \toprule
        $R$& $P$& \emph{Novel} mIoU& \emph{Base} mIoU& mIoU & Mem. (GB)\\
        \midrule
        \multirow{4}{*}{32}
        & 72& 18.48& 11.66& 15.07& \textbf{17.49}\\
        & 90& \textbf{21.16}& \textbf{11.92}& \textbf{16.54}& 17.53\\
        & 120& 20.80& 11.52& 16.16& 17.59\\
        & 180& 17.88& 10.35& 14.11& 17.72\\
        \midrule
        48& 90& 21.08& 11.19& 16.14& 17.58\\
        \bottomrule
    \end{tabular}
    }
\end{table}

\subsubsection{Cylindrical Voxel Resolution} 
To investigate the effect of voxel resolution in the cylindrical representation, we vary the numbers of radial and angular partitions $(R,~P)$, while fixing the vertical dimension at $Z=8$. 
As shown in Tab.~\ref{tab:psm_resolution}, a coarse angular discretization, such as $P=72$, provides insufficient resolution for modeling fine-grained azimuthal structures and results in lower semantic occupancy accuracy.
Increasing the angular resolution to $P=120$ or $P=180$ yields a denser cylindrical representation but does not provide consistent performance gains, while increasing training memory consumption.
Among the evaluated configurations, $(R,P,Z)=(32,90,8)$ achieves the best trade-off between segmentation accuracy and efficiency, yielding $21.16$ \emph{Novel} mIoU, $11.92$ \emph{Base} mIoU, and $16.54$ overall mIoU with moderate memory usage ($17.53$ GB) during training.
Therefore, this setting is adopted as the configuration for our model.

\subsubsection{Occupancy Cost Aggregation} 
As shown in Tab.~\ref{tab:oca}, we further validate the contribution of each component in the proposed OCA. Starting from the setting without cost aggregation (\textbf{A}), introducing the basic cost aggregation branch (\textbf{B}) yields a clear improvement, confirming that modeling voxel–text similarity benefits open-vocabulary occupancy reasoning. Further incorporating spatial aggregation (\textbf{C}) improves structural consistency among voxels, while class aggregation (\textbf{D}) effectively captures inter-class dependencies in the joint embedding space. Finally, integrating embedding guidance (\textbf{E}) achieves the best performance, reaching $21.16$ \emph{Novel} mIoU and $16.54$ overall mIoU. 
Notably, compared with configuration \textbf{A}, the complete configuration (\textbf{E}) yields significant gains, especially on novel categories, with a $+2.57$ mIoU improvement on novel classes and $+1.72$ mIoU in overall performance. 
These consistent improvements indicate that progressively refined cost aggregation and cross-modal guidance substantially enhance both semantic alignment accuracy and generalization to unseen novel categories.

\begin{table}[pos=!t]
    \centering
    \caption{\textbf{Ablation study of OCA.} We conduct an ablation study by progressively adding components to the aggregation branch.
    NMA is not applicable to configuration $\textbf{A}$ because the required cost aggregation branch is absent, and is enabled in configurations $\textbf{B}{-}\textbf{E}$.
    }
    \label{tab:oca}
    \resizebox{\linewidth}{!}{%
    \begin{tabular}{@{}ll|c|cc|c@{}}
        \toprule
        & Components & NMA & \emph{Novel} mIoU & \emph{Base} mIoU & mIoU \\
        \midrule
        \textbf{A} & \textit{w/o} cost agg.& & 18.59& 11.05& 14.82\\
        \midrule
        \textbf{B} & cost agg. & \multirow{4}{*}{\ding{51}}& 18.84& 11.36& 15.10\\
        \textbf{C} & \textbf{B} + spatial agg. & & 18.99& 11.33& 15.16\\
        \textbf{D} & \textbf{C} + class agg. & & 19.67& 11.13& 15.40\\
        \textbf{E} & \textbf{D} + embedding guidance & & \textbf{21.16}& \textbf{11.92}& \textbf{16.54}\\
        \bottomrule
    \end{tabular}
    }
\end{table}

\begin{figure}[pos=!t]
    \centering
    \includegraphics[width=.7\linewidth]{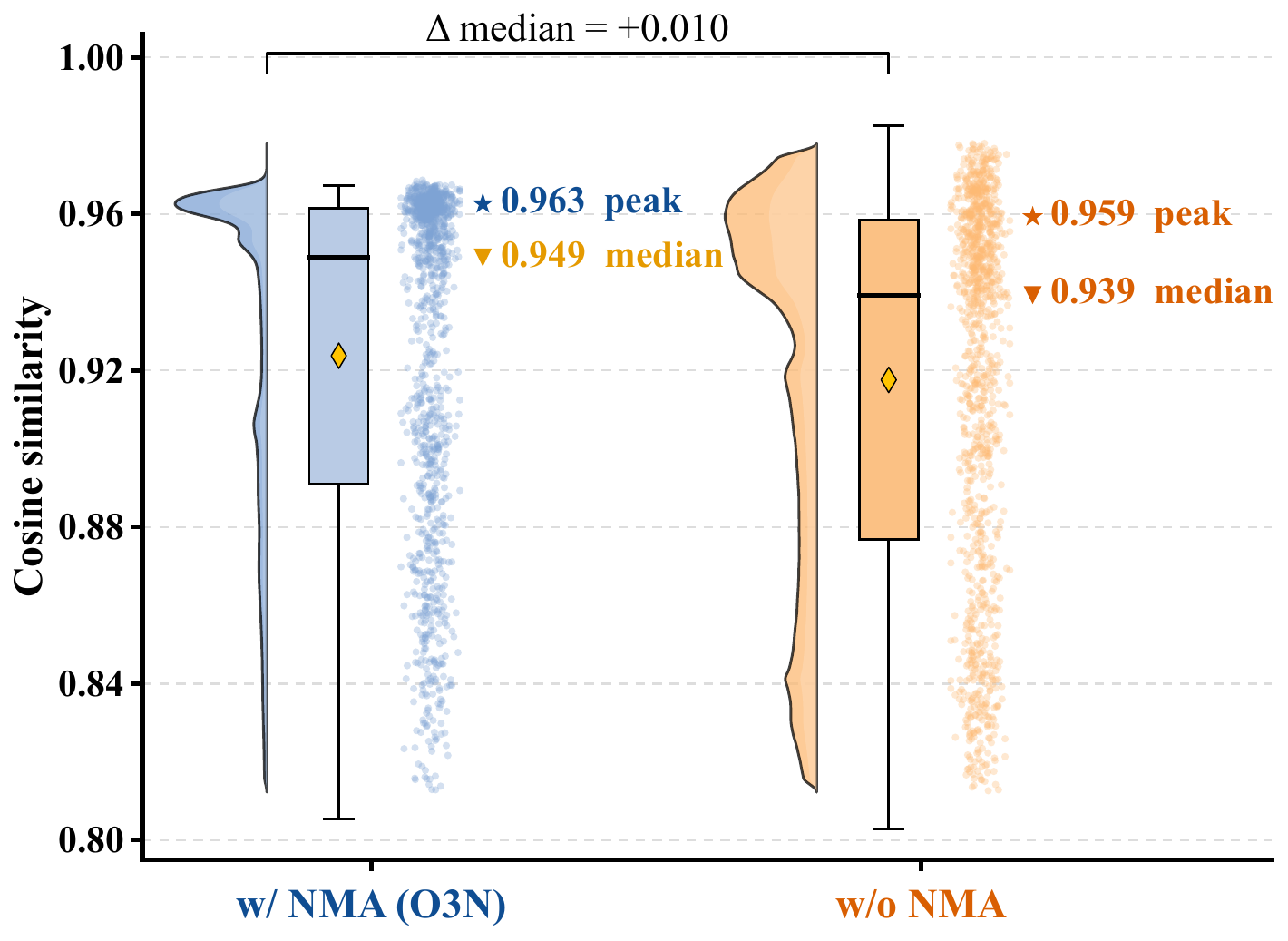}
    \caption{\textbf{Distribution of similarity scores between voxels and text embeddings} with and without NMA during inference.}
    \label{fig:similarityScores}
\end{figure}

\begin{figure}[pos=!t]
    \centering
    \includegraphics[width=.6\linewidth]{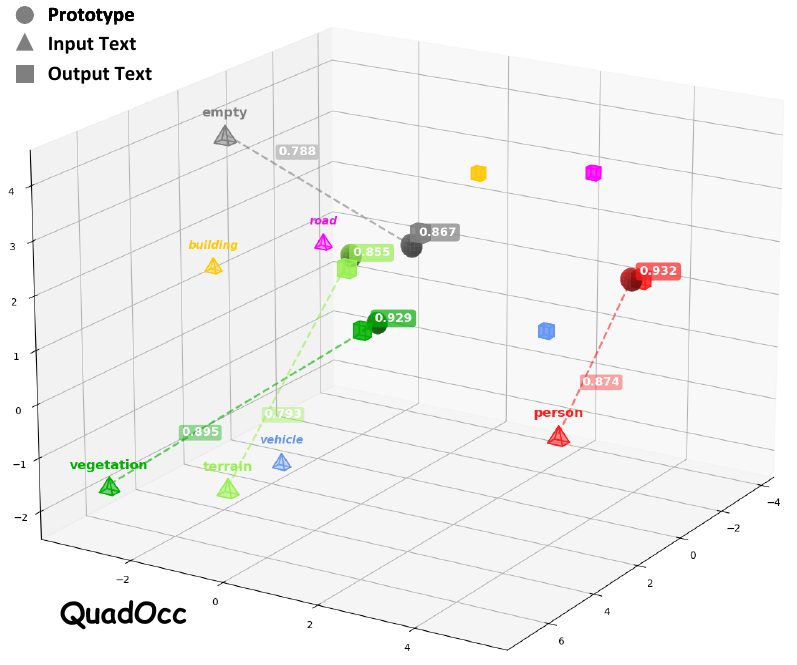}
    \caption{\textbf{PCA visualization} of prototypes and text embeddings on the QuadOcc~\citep{shi2026oneocc} dataset.}
    \label{fig:pca_prototypes}
\end{figure}

\subsubsection{Effectiveness of NMA} 
Before constructing the omnidirectional geometry-semantic consistency cost volume, NMA introduces a gradient-free alignment mechanism that mitigates interference caused by modality gaps or noise in vision-language model predictions. Crucially, it decouples the optimization of the text branch, enabling the construction of a consistent and generalizable \textit{``$\mathbf{P}$-$\mathbf{V}$-$\mathbf{T}$''} representational triplet. To further elucidate its impact, Fig.~\ref{fig:similarityScores} shows the distribution of similarity scores between voxels and text embeddings with and without NMA during inference. 
O3N demonstrates a notably tighter and more stable distribution, with the majority of samples concentrated within the narrow range of $0.94{\sim}0.97$. 
In contrast, the \emph{w/o} NMA variant shows a substantially broader distribution, with a non-negligible fraction of low-similarity samples, suggesting that its cross-modal alignment is more susceptible to noise and exhibits weaker generalization. 

Furthermore, Fig.~\ref{fig:pca_prototypes} presents PCA visualizations of the pixel prototypes and text embeddings before and after applying NMA.
The visualization shows that NMA shifts the text embeddings toward their corresponding semantic prototypes in the projected latent space, indicating a reduced cross-modal representation discrepancy.
To complement this qualitative observation, we compute the cosine similarities between the prototypes and text embeddings in the original embedding space.
After applying NMA, the similarity increases from $0.874$ to $0.932$ for \emph{person} and from $0.793$ to $0.855$ for \emph{terrain}.
Together with the improvements in \emph{Novel} mIoU, these observations further corroborate the effectiveness of NMA in enhancing semantic generalization capability in omnidirectional open-vocabulary occupancy.

\begin{figure}[pos=!t]
    \centering
    \includegraphics[width=.8\linewidth]{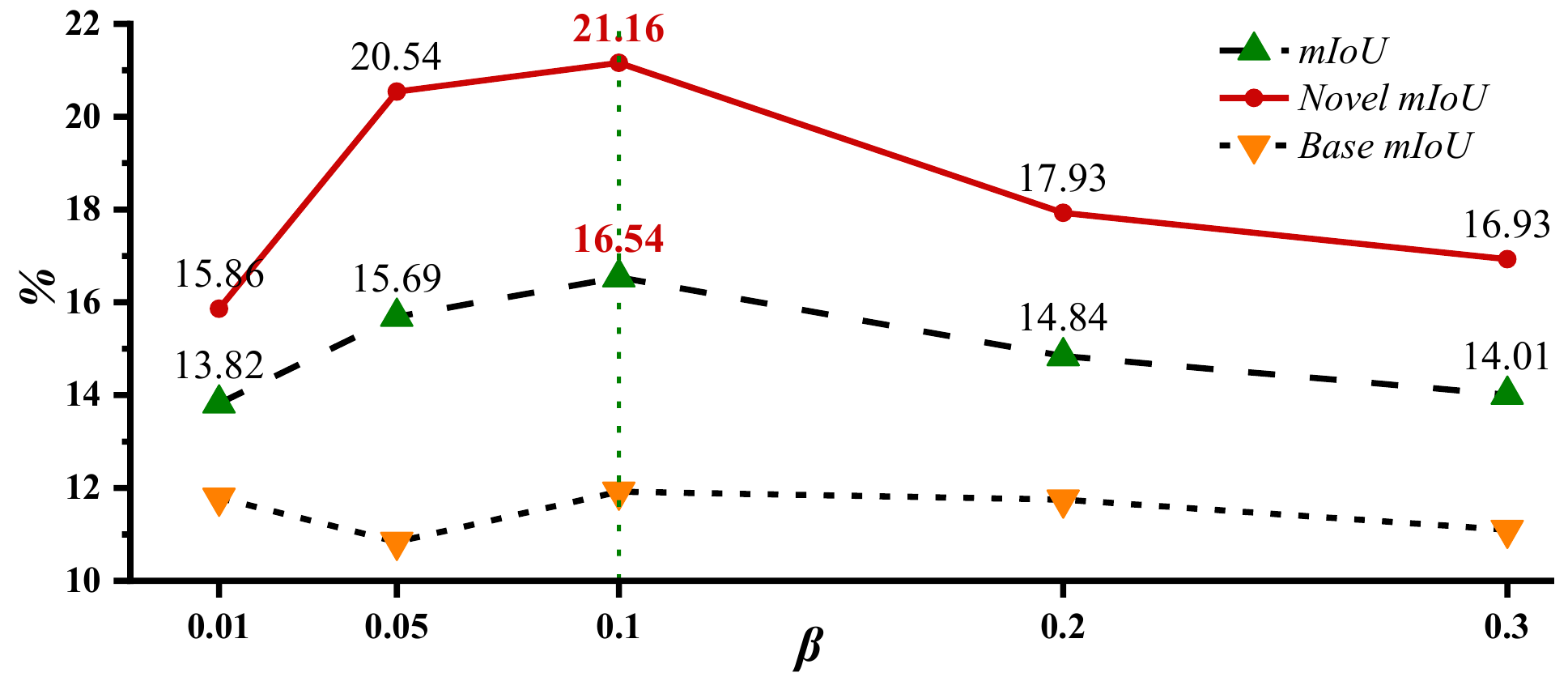}
    \caption{\textbf{Analysis of walking probability $\beta$ in NMA learning}: impact on cross-modal alignment and semantic generalization.}
    \label{fig:nmaBeta}
\end{figure}

\subsubsection{Probability of Walking} 
We further investigate the impact of the walking probability $\beta$ of the NMA in Fig.~\ref{fig:nmaBeta}. As $\beta$ increases from $0.01$ to $0.1$, the model performance consistently improves and reaches its global peak at $0.1$, achieving $21.16$ \emph{Novel} mIoU and $16.54$ overall mIoU. This trend validates the effectiveness of our proposed gradient-free modality alignment mechanism: a moderate walking probability encourages flexible exploration and adaptive fusion among \textit{``pixel–voxel–text''} representations, thereby markedly enhancing cross-modal semantic consistency and generalization to unseen novel categories. In contrast, when $\beta$ is too large (\ie, $0.2$ or $0.3$), excessive emphasis on affinity-based propagation destabilizes the alignment process, leading to severe performance degradation.

\begin{figure}[pos=!t]
    \centering
    \begin{subfigure}{\linewidth}
        \centering
        \includegraphics[width=.9\linewidth]{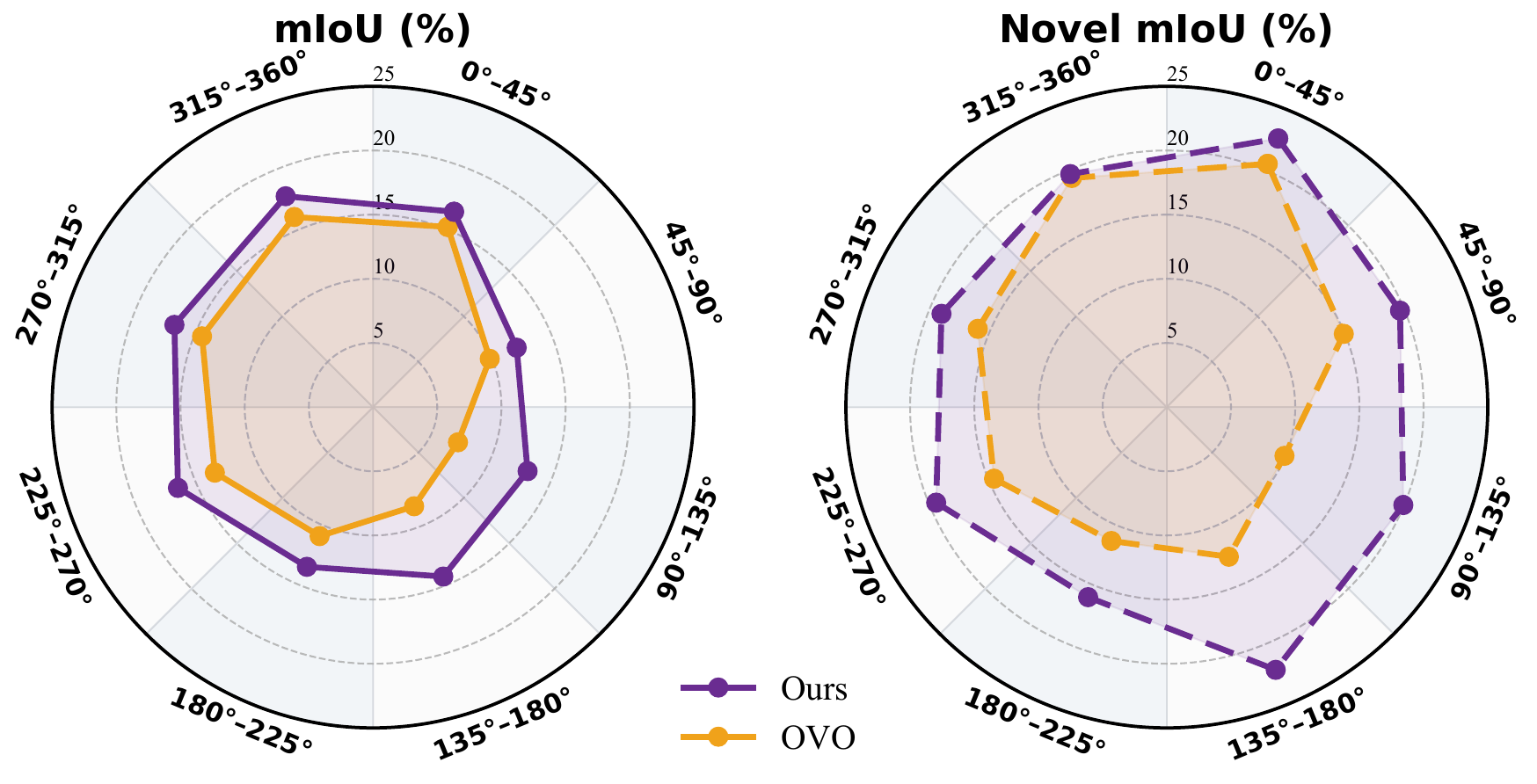}
        \caption{Azimuth-wise regional evaluation.}
        \label{fig:region_fov_miou}
    \end{subfigure}
    
    \begin{subfigure}{\linewidth}
        \centering
        \includegraphics[width=.9\linewidth]{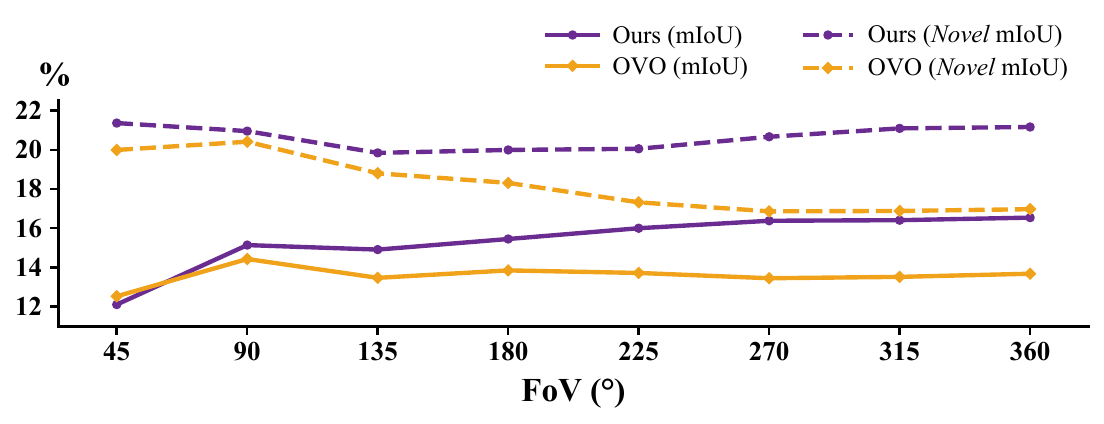}
        \caption{Robustness to different input FoVs.}
        \label{fig:visible_fov_miou}
    \end{subfigure}
    \caption{
    \textbf{Performance analysis using different FoV inputs.} 
    O3N and OVO~\citep{tan2023ovo} are trained once with $360^{\circ}$
    inputs and evaluated without retraining.
    (a) Regional mIoU across eight $45^{\circ}$ azimuth sectors, with $0^{\circ}$ aligned to the positive $x$-axis.
    (b) Visible-region mIoU under forward-centered input FoVs from $45^{\circ}$ to $360^{\circ}$.
    }
    \label{fig:fov}
\end{figure}

\subsubsection{Analysis under Different Fields of View}
We analyze FoV robustness from two complementary perspectives. First, using the complete $360^{\circ}$ input, the surrounding voxel space is divided into eight azimuthal sectors, and each sector is assessed independently. As illustrated in Fig.~\ref{fig:fov}, O3N consistently improves both overall and \emph{novel} mIoU across all spatial directions, with particularly clear gains in regions outside the forward-facing view. This demonstrates that O3N effectively propagates visual and semantic information across the surrounding space, leading to more spatially balanced omnidirectional perception.

Second, we vary the input FoV through forward-centered horizontal cropping without fine-tuning and evaluate only the corresponding visible voxel regions. O3N remains competitive under highly restricted FoVs and exhibits increasingly clear advantages as the visible range expands. In contrast to OVO~\citep{tan2023ovo}, whose performance saturates or degrades with wider inputs, O3N benefits consistently from the additional surrounding context while maintaining stable novel-class recognition. These results indicate that O3N effectively correlates voxel, visual, and textual representations across different spatial ranges, demonstrating robust adaptability to pinhole, fisheye, and panoramic camera configurations.

\section{Conclusion}
\label{sec:conclusion}

In this paper, we investigate omnidirectional open-vocabulary occupancy prediction task and propose O3N, the first framework for open-vocabulary 3D occupancy prediction from a single omnidirectional RGB image, unifying omnidirectional visual perception, spatial geometry, and open-vocabulary semantic prediction.
The proposed Polar-spiral Mamba (PsM) module enables seamless and continuous spatial perception and long-range contextual modeling across 360{\textdegree}.
Furthermore, Occupancy Cost Aggregation (OCA) provides a unified mechanism for geometric and semantic learning, ensuring that the reconstructed geometry is closely aligned with the underlying semantics.
Natural Modality Alignment (NMA) aligns visual features, voxel embeddings, and textual semantics, yielding a consistent and robust \textit{``pixel-voxel-text''} triplet representation.
O3N achieves state-of-the-art open-vocabulary occupancy prediction under omnidirectional visual perception, providing a scalable 3D perception framework for comprehensive and generalizable scene understanding in complex open-world environments.

Despite these advances, O3N remains constrained by the domain gap arising from vision-language models pretrained primarily on pinhole imagery and by the depth ambiguity inherent in monocular perception.
Moreover, recognizing rare novel categories remains challenging, particularly when balancing geometric–semantic consistency with category-level recall under highly imbalanced semantic distributions.
Future work could explore omnidirectional vision-language pretraining to improve semantic representations, together with geometry- and uncertainty-aware projection strategies to enhance spatial reconstruction.
Extending O3N to temporal observations and more diverse real-world environments may further improve prediction consistency and generalization. 
Integrating O3N into downstream intelligent systems for robotic navigation, autonomous exploration, and spatial reasoning should also be explored to evaluate its practical utility in real-world applications.

\ifshowauthors

\section*{CRediT authorship contribution statement}
\textbf{Mengfei Duan:} Conceptualization, Data curation, Formal analysis, Investigation, Methodology, Software, Validation, Visualization, Writing – original draft, Writing – review \& editing.
\textbf{Hao Shi:} Conceptualization, Investigation, Methodology, Validation, Writing – original draft, Writing – review \& editing.
\textbf{Fei Teng:} Data curation, Visualization, Writing – original draft, Writing – review \& editing.
\textbf{Guoqiang Zhao:} Data curation, Writing – original draft, Writing – review \& editing.
\textbf{Yuheng Zhang:} Data curation, Writing – review \& editing.
\textbf{Zhiyong Li:} Funding acquisition, Project administration, Resources, Supervision.
\textbf{Kailun Yang:} Conceptualization, Funding acquisition, Project administration, Resources, Supervision, Writing – original draft, Writing – review \& editing.

\section*{Declaration of competing interest}
The authors declare that they have no known competing financial interests or personal relationships that could have appeared to influence the work reported in this paper.

\section*{Acknowledgements}
This work was supported in part by the National Natural Science Foundation of China (Grant No. 62388101, No. U23A20341, and No. 62473139), in part by the Major Scientific and Technological Research Plan of Hunan Province (No. 2025QK1005), in part by the Hunan Provincial Research and Development Project (Grant No. 2025QK3019), and in part by the State Key Laboratory of Autonomous Intelligent Unmanned Systems (the opening project number ZZKF2025-2-10).

\section*{Data availability}
Data will be made available on request.

\fi

\bibliographystyle{cas-model2-names}

\bibliography{bib}

\begin{thebibliography}{103}
\expandafter\ifx\csname natexlab\endcsname\relax\def\natexlab#1{#1}\fi
\providecommand{\url}[1]{\texttt{#1}}
\providecommand{\href}[2]{#2}
\providecommand{\path}[1]{#1}
\providecommand{\DOIprefix}{doi:}
\providecommand{\ArXivprefix}{arXiv:}
\providecommand{\URLprefix}{URL: }
\providecommand{\Pubmedprefix}{pmid:}
\providecommand{\doi}[1]{\href{http://dx.doi.org/#1}{\path{#1}}}
\providecommand{\Pubmed}[1]{\href{pmid:#1}{\path{#1}}}
\providecommand{\bibinfo}[2]{#2}
\ifx\xfnm\relax \def\xfnm[#1]{\unskip,\space#1}\fi
\bibitem[{Ai et~al.(2025)Ai, Cao and Wang}]{ai2025survey}
\bibinfo{author}{Ai, H.}, \bibinfo{author}{Cao, Z.}, \bibinfo{author}{Wang, L.}, \bibinfo{year}{2025}.
\newblock \bibinfo{title}{A survey of representation learning, optimization strategies, and applications for omnidirectional vision}.
\newblock \bibinfo{journal}{International Journal of Computer Vision} \bibinfo{volume}{133}, \bibinfo{pages}{4973--5012}.
\bibitem[{Boeder et~al.(2025)Boeder, Gigengack and Risse}]{boeder2025langocc}
\bibinfo{author}{Boeder, S.}, \bibinfo{author}{Gigengack, F.}, \bibinfo{author}{Risse, B.}, \bibinfo{year}{2025}.
\newblock \bibinfo{title}{{LangOcc:} {Open} vocabulary occupancy estimation via volume rendering}, in: \bibinfo{booktitle}{3DV}, pp. \bibinfo{pages}{200--210}.
\bibitem[{Cai et~al.(2021)Cai, Ravichandran, Maji, Fowlkes, Tu and Soatto}]{cai2021exponential}
\bibinfo{author}{Cai, Z.}, \bibinfo{author}{Ravichandran, A.}, \bibinfo{author}{Maji, S.}, \bibinfo{author}{Fowlkes, C.}, \bibinfo{author}{Tu, Z.}, \bibinfo{author}{Soatto, S.}, \bibinfo{year}{2021}.
\newblock \bibinfo{title}{Exponential moving average normalization for self-supervised and semi-supervised learning}, in: \bibinfo{booktitle}{CVPR}, pp. \bibinfo{pages}{194--203}.
\bibitem[{Cao and de~Charette(2022)}]{cao2022monoscene}
\bibinfo{author}{Cao, A.Q.}, \bibinfo{author}{de~Charette, R.}, \bibinfo{year}{2022}.
\newblock \bibinfo{title}{{MonoScene:} {Monocular} {3D} semantic scene completion}, in: \bibinfo{booktitle}{CVPR}, pp. \bibinfo{pages}{3981--3991}.
\bibitem[{Cao and Vu(2026)}]{cao2026occany}
\bibinfo{author}{Cao, A.Q.}, \bibinfo{author}{Vu, T.H.}, \bibinfo{year}{2026}.
\newblock \bibinfo{title}{{OccAny:} {Generalized} unconstrained urban {3D} occupancy}, in: \bibinfo{booktitle}{CVPR}, pp. \bibinfo{pages}{28599--28609}.
\bibitem[{Cao~Dinh et~al.(2024)Cao~Dinh, Kim and Cho}]{cao2024geometric}
\bibinfo{author}{Cao~Dinh, D.}, \bibinfo{author}{Kim, S.J.}, \bibinfo{author}{Cho, K.}, \bibinfo{year}{2024}.
\newblock \bibinfo{title}{Geometric exploitation for indoor panoramic semantic segmentation}, in: \bibinfo{booktitle}{NeurIPS}, pp. \bibinfo{pages}{26355--26376}.
\bibitem[{Chang et~al.(2026)Chang, Cao, Zheng, Mi and Dong}]{chang2026denoise}
\bibinfo{author}{Chang, Y.}, \bibinfo{author}{Cao, Z.}, \bibinfo{author}{Zheng, X.}, \bibinfo{author}{Mi, X.}, \bibinfo{author}{Dong, Z.}, \bibinfo{year}{2026}.
\newblock \bibinfo{title}{Denoise and align: {Towards} source-free {UDA} for robust panoramic semantic segmentation}, in: \bibinfo{booktitle}{CVPR}, pp. \bibinfo{pages}{32226--32235}.
\bibitem[{Chen et~al.(2018)Chen, Zhu, Papandreou, Schroff and Adam}]{chen2018deeplabv3+}
\bibinfo{author}{Chen, L.C.}, \bibinfo{author}{Zhu, Y.}, \bibinfo{author}{Papandreou, G.}, \bibinfo{author}{Schroff, F.}, \bibinfo{author}{Adam, H.}, \bibinfo{year}{2018}.
\newblock \bibinfo{title}{Encoder-decoder with atrous separable convolution for semantic image segmentation}, in: \bibinfo{booktitle}{ECCV}, pp. \bibinfo{pages}{833--851}.
\bibitem[{Cho et~al.(2024)Cho, Shin, Hong, Arnab, Seo and Kim}]{cho2024cat_seg}
\bibinfo{author}{Cho, S.}, \bibinfo{author}{Shin, H.}, \bibinfo{author}{Hong, S.}, \bibinfo{author}{Arnab, A.}, \bibinfo{author}{Seo, P.H.}, \bibinfo{author}{Kim, S.}, \bibinfo{year}{2024}.
\newblock \bibinfo{title}{{CAT-Seg:} {Cost} aggregation for open-vocabulary semantic segmentation}, in: \bibinfo{booktitle}{CVPR}, pp. \bibinfo{pages}{4113--4123}.
\bibitem[{Duan et~al.(2025)}]{duan2025sdgocc}
\bibinfo{author}{Duan, Z.}, et~al., \bibinfo{year}{2025}.
\newblock \bibinfo{title}{{SDGOCC:} {Semantic} and depth-guided bird's-eye view transformation for {3D} multimodal occupancy prediction}, in: \bibinfo{booktitle}{CVPR}, pp. \bibinfo{pages}{6751--6760}.
\bibitem[{E et~al.(2025)}]{wenke2025dur360bev}
\bibinfo{author}{E, W.}, et~al., \bibinfo{year}{2025}.
\newblock \bibinfo{title}{{Dur360BEV:} {A} real-world 360-degree single camera dataset and benchmark for bird-eye view mapping in autonomous driving}, in: \bibinfo{booktitle}{ICRA}, pp. \bibinfo{pages}{3737--3744}.
\bibitem[{Gao et~al.(2022)Gao, Yang, Shi, Wang and Bai}]{gao2022review}
\bibinfo{author}{Gao, S.}, \bibinfo{author}{Yang, K.}, \bibinfo{author}{Shi, H.}, \bibinfo{author}{Wang, K.}, \bibinfo{author}{Bai, J.}, \bibinfo{year}{2022}.
\newblock \bibinfo{title}{Review on panoramic imaging and its applications in scene understanding}.
\newblock \bibinfo{journal}{IEEE Transactions on Instrumentation and Measurement} \bibinfo{volume}{71}, \bibinfo{pages}{1--34}.
\bibitem[{Gao et~al.(2025)Gao, Xiang, Zhong and Wang}]{gao2025loc}
\bibinfo{author}{Gao, Y.}, \bibinfo{author}{Xiang, X.}, \bibinfo{author}{Zhong, S.}, \bibinfo{author}{Wang, G.}, \bibinfo{year}{2025}.
\newblock \bibinfo{title}{{LOC:} {A} general language-guided framework for open-set {3D} occupancy prediction}.
\newblock \bibinfo{journal}{arXiv preprint arXiv:2510.22141} .
\bibitem[{Grady(2006)}]{grady2006random}
\bibinfo{author}{Grady, L.}, \bibinfo{year}{2006}.
\newblock \bibinfo{title}{Random walks for image segmentation}.
\newblock \bibinfo{journal}{IEEE Transactions on Pattern Analysis and Machine Intelligence} \bibinfo{volume}{28}, \bibinfo{pages}{1768--1783}.
\bibitem[{Guo et~al.(2025)Guo, Li and Han}]{guo2025overlapocc}
\bibinfo{author}{Guo, S.}, \bibinfo{author}{Li, J.}, \bibinfo{author}{Han, S.}, \bibinfo{year}{2025}.
\newblock \bibinfo{title}{Overlapocc: Leveraging overlap regions of surround-view cameras for 3d semantic occupancy prediction}.
\newblock \bibinfo{journal}{Expert Systems with Applications} \bibinfo{volume}{293}, \bibinfo{pages}{128701}.
\bibitem[{Guttikonda and Rambach(2024)}]{guttikonda2024single}
\bibinfo{author}{Guttikonda, S.}, \bibinfo{author}{Rambach, J.}, \bibinfo{year}{2024}.
\newblock \bibinfo{title}{Single frame semantic segmentation using multi-modal spherical images}, in: \bibinfo{booktitle}{WACV}, pp. \bibinfo{pages}{3210--3219}.
\bibitem[{Hu et~al.(2022)Hu, An, Shao and Hu}]{hu2022distortion}
\bibinfo{author}{Hu, X.}, \bibinfo{author}{An, Y.}, \bibinfo{author}{Shao, C.}, \bibinfo{author}{Hu, H.}, \bibinfo{year}{2022}.
\newblock \bibinfo{title}{Distortion convolution module for semantic segmentation of panoramic images based on the image-forming principle}.
\newblock \bibinfo{journal}{IEEE Transactions on Instrumentation and Measurement} \bibinfo{volume}{71}, \bibinfo{pages}{1--12}.
\bibitem[{Huang et~al.(2024a)Huang, Zheng, Zhang, Zhou and Lu}]{huang2024selfocc}
\bibinfo{author}{Huang, Y.}, \bibinfo{author}{Zheng, W.}, \bibinfo{author}{Zhang, B.}, \bibinfo{author}{Zhou, J.}, \bibinfo{author}{Lu, J.}, \bibinfo{year}{2024}a.
\newblock \bibinfo{title}{{SelfOcc:} {Self-supervised} vision-based {3D} occupancy prediction}, in: \bibinfo{booktitle}{CVPR}, pp. \bibinfo{pages}{19946--19956}.
\bibitem[{Huang et~al.(2023)Huang, Zheng, Zhang, Zhou and Lu}]{huang2023tpvformer}
\bibinfo{author}{Huang, Y.}, \bibinfo{author}{Zheng, W.}, \bibinfo{author}{Zhang, Y.}, \bibinfo{author}{Zhou, J.}, \bibinfo{author}{Lu, J.}, \bibinfo{year}{2023}.
\newblock \bibinfo{title}{Tri-perspective view for vision-based {3D} semantic occupancy prediction}, in: \bibinfo{booktitle}{CVPR}, pp. \bibinfo{pages}{9223--9232}.
\bibitem[{Huang et~al.(2024b)Huang, Zheng, Zhang, Zhou and Lu}]{huang2024gaussianformer}
\bibinfo{author}{Huang, Y.}, \bibinfo{author}{Zheng, W.}, \bibinfo{author}{Zhang, Y.}, \bibinfo{author}{Zhou, J.}, \bibinfo{author}{Lu, J.}, \bibinfo{year}{2024}b.
\newblock \bibinfo{title}{{GaussianFormer:} {Scene} as gaussians for vision-based {3D} semantic occupancy prediction}, in: \bibinfo{booktitle}{ECCV}, pp. \bibinfo{pages}{376--393}.
\bibitem[{Jiang et~al.(2024)}]{jiang2024openocc}
\bibinfo{author}{Jiang, H.}, et~al., \bibinfo{year}{2024}.
\newblock \bibinfo{title}{{OpenOcc:} {Open} vocabulary {3D} scene reconstruction via occupancy representation}, in: \bibinfo{booktitle}{IROS}, pp. \bibinfo{pages}{1145--1152}.
\bibitem[{Jiang et~al.(2026a)Jiang, Zhao, Zhu, Li, Chen and Yao}]{jiang2026augmenting}
\bibinfo{author}{Jiang, J.}, \bibinfo{author}{Zhao, S.}, \bibinfo{author}{Zhu, J.}, \bibinfo{author}{Li, M.}, \bibinfo{author}{Chen, X.}, \bibinfo{author}{Yao, H.}, \bibinfo{year}{2026}a.
\newblock \bibinfo{title}{Augmenting and contrasting distortion for open panoramic segmentation}.
\newblock \bibinfo{journal}{Science China Information Sciences} .
\bibitem[{Jiang et~al.(2025)Jiang, Zhu, Xu, Chen, Zhao and Yao}]{jiang2025gaussian}
\bibinfo{author}{Jiang, J.}, \bibinfo{author}{Zhu, J.}, \bibinfo{author}{Xu, Z.}, \bibinfo{author}{Chen, X.}, \bibinfo{author}{Zhao, S.}, \bibinfo{author}{Yao, H.}, \bibinfo{year}{2025}.
\newblock \bibinfo{title}{Gaussian constrained diffeomorphic deformation network for panoramic semantic segmentation}, in: \bibinfo{booktitle}{ICASSP}, pp. \bibinfo{pages}{1--5}.
\bibitem[{Jiang et~al.(2026b)Jiang, Zhou, Zuo and Chen}]{jiang2026freeocc}
\bibinfo{author}{Jiang, Z.}, \bibinfo{author}{Zhou, C.}, \bibinfo{author}{Zuo, X.}, \bibinfo{author}{Chen, C.}, \bibinfo{year}{2026}b.
\newblock \bibinfo{title}{{FreeOcc:} {Training-free} embodied open-vocabulary occupancy prediction}, in: \bibinfo{booktitle}{RSS}, pp. \bibinfo{pages}{1--12}.
\bibitem[{Katharopoulos et~al.(2020)Katharopoulos, Vyas, Pappas and Fleuret}]{katharopoulos2020transformers}
\bibinfo{author}{Katharopoulos, A.}, \bibinfo{author}{Vyas, A.}, \bibinfo{author}{Pappas, N.}, \bibinfo{author}{Fleuret, F.}, \bibinfo{year}{2020}.
\newblock \bibinfo{title}{Transformers are {RNNs}: {Fast} autoregressive transformers with linear attention}, in: \bibinfo{booktitle}{ICML}, pp. \bibinfo{pages}{5156--5165}.
\bibitem[{Kim et~al.(2022)Kim, Jeong and Sohn}]{kim2022pasts}
\bibinfo{author}{Kim, J.}, \bibinfo{author}{Jeong, S.}, \bibinfo{author}{Sohn, K.}, \bibinfo{year}{2022}.
\newblock \bibinfo{title}{{PASTS:} {Toward} effective distilling transformer for panoramic semantic segmentation}, in: \bibinfo{booktitle}{ICIP}, pp. \bibinfo{pages}{2881--2885}.
\bibitem[{Lan et~al.(2025)Lan, Yang, Xu, Jiang and Wang}]{lan2025deformable}
\bibinfo{author}{Lan, B.}, \bibinfo{author}{Yang, L.}, \bibinfo{author}{Xu, M.}, \bibinfo{author}{Jiang, L.}, \bibinfo{author}{Wang, Y.}, \bibinfo{year}{2025}.
\newblock \bibinfo{title}{Deformable spherical geometry transformer for panoramic semantic segmentation}, in: \bibinfo{booktitle}{ICIP}, pp. \bibinfo{pages}{271--276}.
\bibitem[{Li et~al.(2022)Li, Weinberger, Belongie, Koltun and Ranftl}]{li2022language}
\bibinfo{author}{Li, B.}, \bibinfo{author}{Weinberger, K.Q.}, \bibinfo{author}{Belongie, S.J.}, \bibinfo{author}{Koltun, V.}, \bibinfo{author}{Ranftl, R.}, \bibinfo{year}{2022}.
\newblock \bibinfo{title}{Language-driven semantic segmentation}, in: \bibinfo{booktitle}{ICLR}, pp. \bibinfo{pages}{1--13}.
\bibitem[{Li et~al.(2025a)}]{li2025ago}
\bibinfo{author}{Li, P.}, et~al., \bibinfo{year}{2025}a.
\newblock \bibinfo{title}{{AGO:} {Adaptive} grounding for open world {3D} occupancy prediction}, in: \bibinfo{booktitle}{ICCV}, pp. \bibinfo{pages}{8645--8655}.
\bibitem[{Li et~al.(2026)Li, Bao, Chen and Li}]{li2026rel_sf4pass}
\bibinfo{author}{Li, X.}, \bibinfo{author}{Bao, X.}, \bibinfo{author}{Chen, Z.}, \bibinfo{author}{Li, X.}, \bibinfo{year}{2026}.
\newblock \bibinfo{title}{{REL-SF4PASS:} {Panoramic} semantic segmentation with {REL} depth representation and spherical fusion}, in: \bibinfo{booktitle}{Proc. CVPR}, pp. \bibinfo{pages}{27676--27685}.
\bibitem[{Li et~al.(2023a)Li, Wu, Qi, Wang, Shan and Li}]{li2023sgat4pass}
\bibinfo{author}{Li, X.}, \bibinfo{author}{Wu, T.}, \bibinfo{author}{Qi, Z.}, \bibinfo{author}{Wang, G.}, \bibinfo{author}{Shan, Y.}, \bibinfo{author}{Li, X.}, \bibinfo{year}{2023}a.
\newblock \bibinfo{title}{{SGAT4PASS:} {Spherical} geometry-aware transformer for panoramic semantic segmentation}, in: \bibinfo{booktitle}{IJCAI}, pp. \bibinfo{pages}{1125--1133}.
\bibitem[{Li et~al.(2023b)}]{li2023voxformer}
\bibinfo{author}{Li, Y.}, et~al., \bibinfo{year}{2023}b.
\newblock \bibinfo{title}{{VoxFormer:} {Sparse} voxel transformer for camera-based {3D} semantic scene completion}, in: \bibinfo{booktitle}{CVPR}, pp. \bibinfo{pages}{9087--9098}.
\bibitem[{Li et~al.(2025b)}]{li2024bevformer}
\bibinfo{author}{Li, Z.}, et~al., \bibinfo{year}{2025}b.
\newblock \bibinfo{title}{{BEVFormer:} {Learning} bird's-eye-view representation from {LiDAR-camera} via spatiotemporal transformers}.
\newblock \bibinfo{journal}{IEEE Transactions on Pattern Analysis and Machine Intelligence} \bibinfo{volume}{47}, \bibinfo{pages}{2020--2036}.
\bibitem[{Lin and You(2026)}]{lin2026stg}
\bibinfo{author}{Lin, H.Y.}, \bibinfo{author}{You, T.T.}, \bibinfo{year}{2026}.
\newblock \bibinfo{title}{Stg-nav: Incorporating stg with llm-based semantic reasoning for zero-shot object goal navigation}.
\newblock \bibinfo{journal}{Expert Systems with Applications} , \bibinfo{pages}{132309}.
\bibitem[{Lin et~al.(2025)}]{zheng2025one_flight}
\bibinfo{author}{Lin, X.}, et~al., \bibinfo{year}{2025}.
\newblock \bibinfo{title}{One flight over the gap: A survey from perspective to panoramic vision}.
\newblock \bibinfo{journal}{arXiv preprint arXiv:2509.04444} .
\bibitem[{Liu and Wang(2025)}]{liu2025articubevseg}
\bibinfo{author}{Liu, W.}, \bibinfo{author}{Wang, W.}, \bibinfo{year}{2025}.
\newblock \bibinfo{title}{{ArticuBEVSeg:} {Road} semantic understanding and its application in bird's eye view from panoramic vision system of long combination vehicles}.
\newblock \bibinfo{journal}{IEEE Robotics and Automation Letters} \bibinfo{volume}{10}, \bibinfo{pages}{6864--6871}.
\bibitem[{Lu et~al.(2025)Lu, Yuan, Yang, Liu and Sun}]{lu2025gsff}
\bibinfo{author}{Lu, Z.}, \bibinfo{author}{Yuan, X.}, \bibinfo{author}{Yang, S.}, \bibinfo{author}{Liu, J.}, \bibinfo{author}{Sun, C.}, \bibinfo{year}{2025}.
\newblock \bibinfo{title}{Gsff-slam: 3d semantic gaussian splatting slam via feature field}.
\newblock \bibinfo{journal}{Expert Systems with Applications} , \bibinfo{pages}{129601}.
\bibitem[{Ma et~al.(2021)Ma, Zhang, Yang, Roitberg and Stiefelhagen}]{ma2021densepass}
\bibinfo{author}{Ma, C.}, \bibinfo{author}{Zhang, J.}, \bibinfo{author}{Yang, K.}, \bibinfo{author}{Roitberg, A.}, \bibinfo{author}{Stiefelhagen, R.}, \bibinfo{year}{2021}.
\newblock \bibinfo{title}{{DensePASS:} {Dense} panoramic semantic segmentation via unsupervised domain adaptation with attention-augmented context exchange}, in: \bibinfo{booktitle}{ITSC}, pp. \bibinfo{pages}{2766--2772}.
\bibitem[{Ma et~al.(2024)Ma, Tan, Qu, Ma, Zhang and Xie}]{ma2024cotr}
\bibinfo{author}{Ma, Q.}, \bibinfo{author}{Tan, X.}, \bibinfo{author}{Qu, Y.}, \bibinfo{author}{Ma, L.}, \bibinfo{author}{Zhang, Z.}, \bibinfo{author}{Xie, Y.}, \bibinfo{year}{2024}.
\newblock \bibinfo{title}{{COTR:} {Compact} occupancy transformer for vision-based {3D} occupancy prediction}, in: \bibinfo{booktitle}{CVPR}, pp. \bibinfo{pages}{19936--19945}.
\bibitem[{Mei et~al.(2024)}]{mei2024sgn}
\bibinfo{author}{Mei, J.}, et~al., \bibinfo{year}{2024}.
\newblock \bibinfo{title}{Camera-based {3D} semantic scene completion with sparse guidance network}.
\newblock \bibinfo{journal}{IEEE Transactions on Image Processing} \bibinfo{volume}{33}, \bibinfo{pages}{5468--5481}.
\bibitem[{Meyer(2023)}]{meyer2023matrix}
\bibinfo{author}{Meyer, C.D.}, \bibinfo{year}{2023}.
\newblock \bibinfo{title}{Matrix analysis and applied linear algebra}.
\newblock \bibinfo{publisher}{SIAM}.
\bibitem[{Ming et~al.(2025)}]{ming2025occcylindrical}
\bibinfo{author}{Ming, Z.}, et~al., \bibinfo{year}{2025}.
\newblock \bibinfo{title}{{OccCylindrical:} {Multi-modal} fusion with cylindrical representation for {3D} semantic occupancy prediction}, in: \bibinfo{booktitle}{ITSC}, pp. \bibinfo{pages}{1981--1988}.
\bibitem[{Oh et~al.(2025)}]{oh2025_3d_prototype}
\bibinfo{author}{Oh, G.}, et~al., \bibinfo{year}{2025}.
\newblock \bibinfo{title}{{3D} occupancy prediction with low-resolution queries via prototype-aware view transformation}, in: \bibinfo{booktitle}{CVPR}, pp. \bibinfo{pages}{17134--17144}.
\bibitem[{Orhan and Bastanlar(2022)}]{orhan2022semantic}
\bibinfo{author}{Orhan, S.}, \bibinfo{author}{Bastanlar, Y.}, \bibinfo{year}{2022}.
\newblock \bibinfo{title}{Semantic segmentation of outdoor panoramic images}.
\newblock \bibinfo{journal}{Signal, Image and Video Processing} \bibinfo{volume}{16}, \bibinfo{pages}{643--650}.
\bibitem[{Pan et~al.(2024)}]{pan2024generocc}
\bibinfo{author}{Pan, X.}, et~al., \bibinfo{year}{2024}.
\newblock \bibinfo{title}{{GenerOcc:} {Self-supervised} framework of real-time {3D} occupancy prediction for monocular generic cameras}, in: \bibinfo{booktitle}{IROS}, pp. \bibinfo{pages}{8868--8874}.
\bibitem[{Philion and Fidler(2020)}]{philion2020lift}
\bibinfo{author}{Philion, J.}, \bibinfo{author}{Fidler, S.}, \bibinfo{year}{2020}.
\newblock \bibinfo{title}{Lift, splat, shoot: Encoding images from arbitrary camera rigs by implicitly unprojecting to {3D}}, in: \bibinfo{booktitle}{ECCV}, pp. \bibinfo{pages}{194--210}.
\bibitem[{Radford et~al.(2021)}]{radford2021learning}
\bibinfo{author}{Radford, A.}, et~al., \bibinfo{year}{2021}.
\newblock \bibinfo{title}{Learning transferable visual models from natural language supervision}, in: \bibinfo{booktitle}{ICML}, pp. \bibinfo{pages}{8748--8763}.
\bibitem[{Rold{\~{a}}o et~al.(2020)Rold{\~{a}}o, de~Charette and Verroust{-}Blondet}]{roldao2020lmscnet}
\bibinfo{author}{Rold{\~{a}}o, L.}, \bibinfo{author}{de~Charette, R.}, \bibinfo{author}{Verroust{-}Blondet, A.}, \bibinfo{year}{2020}.
\newblock \bibinfo{title}{{LMSCNet:} {Lightweight} multiscale {3D} semantic completion}, in: \bibinfo{booktitle}{3DV}, pp. \bibinfo{pages}{111--119}.
\bibitem[{Samani et~al.(2023)Samani, Tao, Reddy, Ding and Banerjee}]{samani2023f2bev}
\bibinfo{author}{Samani, E.U.}, \bibinfo{author}{Tao, F.}, \bibinfo{author}{Reddy, D.H.}, \bibinfo{author}{Ding, S.}, \bibinfo{author}{Banerjee, A.G.}, \bibinfo{year}{2023}.
\newblock \bibinfo{title}{{F2BEV:} {Bird's} eye view generation from surround-view fisheye camera images for automated driving}, in: \bibinfo{booktitle}{IROS}, pp. \bibinfo{pages}{9367--9374}.
\bibitem[{Shi et~al.(2026)}]{shi2026oneocc}
\bibinfo{author}{Shi, H.}, et~al., \bibinfo{year}{2026}.
\newblock \bibinfo{title}{{OneOcc:} {Semantic} occupancy prediction for legged robots with a single panoramic camera}, in: \bibinfo{booktitle}{CVPR}, pp. \bibinfo{pages}{14229--14240}.
\bibitem[{Shi et~al.(2024)Shi, Cheng, Zhang, Liu and Wang}]{shi2024occupancy_set_points}
\bibinfo{author}{Shi, Y.}, \bibinfo{author}{Cheng, T.}, \bibinfo{author}{Zhang, Q.}, \bibinfo{author}{Liu, W.}, \bibinfo{author}{Wang, X.}, \bibinfo{year}{2024}.
\newblock \bibinfo{title}{Occupancy as set of points}, in: \bibinfo{booktitle}{ECCV}, pp. \bibinfo{pages}{72--87}.
\bibitem[{Silberman et~al.(2012)Silberman, Hoiem, Kohli and Fergus}]{silberman2012indoor}
\bibinfo{author}{Silberman, N.}, \bibinfo{author}{Hoiem, D.}, \bibinfo{author}{Kohli, P.}, \bibinfo{author}{Fergus, R.}, \bibinfo{year}{2012}.
\newblock \bibinfo{title}{Indoor segmentation and support inference from rgbd images}, in: \bibinfo{booktitle}{ECCV}, \bibinfo{organization}{Springer}. pp. \bibinfo{pages}{746--760}.
\bibitem[{Song et~al.(2017)Song, Yu, Zeng, Chang, Savva and Funkhouser}]{song2017semantic}
\bibinfo{author}{Song, S.}, \bibinfo{author}{Yu, F.}, \bibinfo{author}{Zeng, A.}, \bibinfo{author}{Chang, A.X.}, \bibinfo{author}{Savva, M.}, \bibinfo{author}{Funkhouser, T.A.}, \bibinfo{year}{2017}.
\newblock \bibinfo{title}{Semantic scene completion from a single depth image}, in: \bibinfo{booktitle}{CVPR}, pp. \bibinfo{pages}{190--198}.
\bibitem[{Sun et~al.(2025)Sun, Contreras and Ortiz}]{sun2025dynamic_embodied_occupancy}
\bibinfo{author}{Sun, Y.}, \bibinfo{author}{Contreras, J.}, \bibinfo{author}{Ortiz, J.}, \bibinfo{year}{2025}.
\newblock \bibinfo{title}{Dynamic focused masking for autoregressive embodied occupancy prediction}, in: \bibinfo{booktitle}{NeurIPS}, pp. \bibinfo{pages}{161453--161471}.
\bibitem[{Sun et~al.(2026)Sun, Liu, Shum, Atapour-Abarghouei and Breckon}]{sun2026kd360}
\bibinfo{author}{Sun, Y.}, \bibinfo{author}{Liu, J.}, \bibinfo{author}{Shum, H.P.}, \bibinfo{author}{Atapour-Abarghouei, A.}, \bibinfo{author}{Breckon, T.P.}, \bibinfo{year}{2026}.
\newblock \bibinfo{title}{{KD360-VoxelBEV:} {LiDAR} and 360-degree camera cross modality knowledge distillation for bird's-eye-view segmentation}, in: \bibinfo{booktitle}{WACV}, pp. \bibinfo{pages}{3483--3493}.
\bibitem[{Tai et~al.(2026)}]{tai2026open_sam3d}
\bibinfo{author}{Tai, H.}, et~al., \bibinfo{year}{2026}.
\newblock \bibinfo{title}{Open-vocabulary {SAM3D}: {Towards} training-free open-vocabulary {3D} scene understanding}.
\newblock \bibinfo{journal}{IEEE Transactions on Circuits and Systems for Video Technology} \bibinfo{volume}{36}, \bibinfo{pages}{10943--10956}.
\bibitem[{Tan et~al.(2025)Tan, Chen, Cao, Yan, Ma and Dai}]{tan2025dasc}
\bibinfo{author}{Tan, T.}, \bibinfo{author}{Chen, B.}, \bibinfo{author}{Cao, H.}, \bibinfo{author}{Yan, C.}, \bibinfo{author}{Ma, Y.}, \bibinfo{author}{Dai, F.}, \bibinfo{year}{2025}.
\newblock \bibinfo{title}{{DASC-SPT:} {Towards} self-supervised panoramic semantic segmentation}, in: \bibinfo{booktitle}{WACV}, pp. \bibinfo{pages}{8796--8805}.
\bibitem[{Tan et~al.(2023)Tan, Dong, Zhang, Zhang, Ji and Li}]{tan2023ovo}
\bibinfo{author}{Tan, Z.}, \bibinfo{author}{Dong, Z.}, \bibinfo{author}{Zhang, C.}, \bibinfo{author}{Zhang, W.}, \bibinfo{author}{Ji, H.}, \bibinfo{author}{Li, H.}, \bibinfo{year}{2023}.
\newblock \bibinfo{title}{{OVO:} {Open-vocabulary} occupancy}.
\newblock \bibinfo{journal}{arXiv preprint arXiv:2305.16133} .
\bibitem[{Tang et~al.(2024)}]{tang2024sparseocc}
\bibinfo{author}{Tang, P.}, et~al., \bibinfo{year}{2024}.
\newblock \bibinfo{title}{{SparseOcc:} {Rethinking} sparse latent representation for vision-based semantic occupancy prediction}, in: \bibinfo{booktitle}{CVPR}, pp. \bibinfo{pages}{15035--15044}.
\bibitem[{Teng et~al.(2024)}]{teng2024360bev}
\bibinfo{author}{Teng, Z.}, et~al., \bibinfo{year}{2024}.
\newblock \bibinfo{title}{{360BEV:} {Panoramic} semantic mapping for indoor bird's-eye view}, in: \bibinfo{booktitle}{WACV}, pp. \bibinfo{pages}{372--381}.
\bibitem[{Tian et~al.(2023)}]{tian2023occ3d}
\bibinfo{author}{Tian, X.}, et~al., \bibinfo{year}{2023}.
\newblock \bibinfo{title}{{Occ3D:} {A} large-scale {3D} occupancy prediction benchmark for autonomous driving}, in: \bibinfo{booktitle}{NeurIPS}, pp. \bibinfo{pages}{64318--64330}.
\bibitem[{Tian et~al.(2026)Tian, Bai, Luo, Wang, Zhang, Guo, Lv and Wang}]{tian2026mambaocc}
\bibinfo{author}{Tian, Y.}, \bibinfo{author}{Bai, S.}, \bibinfo{author}{Luo, Z.}, \bibinfo{author}{Wang, Y.}, \bibinfo{author}{Zhang, H.}, \bibinfo{author}{Guo, B.}, \bibinfo{author}{Lv, Y.}, \bibinfo{author}{Wang, F.Y.}, \bibinfo{year}{2026}.
\newblock \bibinfo{title}{Mambaocc: Visual state space models for bev-based occupancy prediction with local adaptive reordering}.
\newblock \bibinfo{journal}{Expert Systems with Applications} , \bibinfo{pages}{131305}.
\bibitem[{Tong et~al.(2023)}]{tong2023scene_as_occupancy}
\bibinfo{author}{Tong, W.}, et~al., \bibinfo{year}{2023}.
\newblock \bibinfo{title}{Scene as occupancy}, in: \bibinfo{booktitle}{ICCV}, pp. \bibinfo{pages}{8372--8381}.
\bibitem[{Vobecky et~al.(2023)}]{vobecky2023pop_3d}
\bibinfo{author}{Vobecky, A.}, et~al., \bibinfo{year}{2023}.
\newblock \bibinfo{title}{{POP-3D:} {Open-vocabulary} {3D} occupancy prediction from images}, in: \bibinfo{booktitle}{NeurIPS}, pp. \bibinfo{pages}{50545--50557}.
\bibitem[{Wang et~al.(2025)}]{wang2025occrwkv}
\bibinfo{author}{Wang, J.}, et~al., \bibinfo{year}{2025}.
\newblock \bibinfo{title}{{OccRWKV:} {Rethinking} efficient {3D} semantic occupancy prediction with linear complexity}, in: \bibinfo{booktitle}{ICRA}, pp. \bibinfo{pages}{10915--10921}.
\bibitem[{Wang et~al.(2024)Wang, Feng and Wan}]{wang2024multi}
\bibinfo{author}{Wang, X.}, \bibinfo{author}{Feng, W.}, \bibinfo{author}{Wan, L.}, \bibinfo{year}{2024}.
\newblock \bibinfo{title}{Multi-modal fusion architecture search for camera-based semantic scene completion}.
\newblock \bibinfo{journal}{Expert Systems with Applications} \bibinfo{volume}{243}, \bibinfo{pages}{122885}.
\bibitem[{Wang et~al.(2023)}]{wang2023openoccupancy}
\bibinfo{author}{Wang, X.}, et~al., \bibinfo{year}{2023}.
\newblock \bibinfo{title}{{OpenOccupancy:} {A} large scale benchmark for surrounding semantic occupancy perception}, in: \bibinfo{booktitle}{ICCV}, pp. \bibinfo{pages}{17804--17813}.
\bibitem[{Wei et~al.(2024)Wei, Zheng, Liu, Hu, Zhang and Stiefelhagen}]{wei2024onebev}
\bibinfo{author}{Wei, J.}, \bibinfo{author}{Zheng, J.}, \bibinfo{author}{Liu, R.}, \bibinfo{author}{Hu, J.}, \bibinfo{author}{Zhang, J.}, \bibinfo{author}{Stiefelhagen, R.}, \bibinfo{year}{2024}.
\newblock \bibinfo{title}{{OneBEV:} {Using} one panoramic image for bird's-eye-view semantic mapping}, in: \bibinfo{booktitle}{ACCV}, pp. \bibinfo{pages}{377--393}.
\bibitem[{Wei et~al.(2023)Wei, Zhao, Zheng, Zhu, Zhou and Lu}]{wei2023surroundocc}
\bibinfo{author}{Wei, Y.}, \bibinfo{author}{Zhao, L.}, \bibinfo{author}{Zheng, W.}, \bibinfo{author}{Zhu, Z.}, \bibinfo{author}{Zhou, J.}, \bibinfo{author}{Lu, J.}, \bibinfo{year}{2023}.
\newblock \bibinfo{title}{{SurroundOcc:} {Multi-camera} {3D} occupancy prediction for autonomous driving}, in: \bibinfo{booktitle}{ICCV}, pp. \bibinfo{pages}{21672--21683}.
\bibitem[{Wu et~al.(2025a)Wu, Li, Cao, Li, Du and Li}]{wu2025omniocc}
\bibinfo{author}{Wu, C.}, \bibinfo{author}{Li, J.}, \bibinfo{author}{Cao, J.}, \bibinfo{author}{Li, M.}, \bibinfo{author}{Du, S.}, \bibinfo{author}{Li, Y.}, \bibinfo{year}{2025}a.
\newblock \bibinfo{title}{{OmniOcc:} {Cylindrical} voxel-based semantic occupancy prediction for omnidirectional vision systems}.
\newblock \bibinfo{journal}{IEEE Access} \bibinfo{volume}{13}, \bibinfo{pages}{139944--139952}.
\bibitem[{Wu et~al.(2025b)Wu, Zheng, Zuo, Huang, Zhou and Lu}]{wu2024embodiedocc}
\bibinfo{author}{Wu, Y.}, \bibinfo{author}{Zheng, W.}, \bibinfo{author}{Zuo, S.}, \bibinfo{author}{Huang, Y.}, \bibinfo{author}{Zhou, J.}, \bibinfo{author}{Lu, J.}, \bibinfo{year}{2025}b.
\newblock \bibinfo{title}{{EmbodiedOcc:} {Embodied} {3D} occupancy prediction for vision-based online scene understanding}, in: \bibinfo{booktitle}{ICCV}, pp. \bibinfo{pages}{26360--26370}.
\bibitem[{Xiao et~al.(2025)Xiao, Li, Zhang, Meng and Zhang}]{xiao2024spatial}
\bibinfo{author}{Xiao, C.}, \bibinfo{author}{Li, M.}, \bibinfo{author}{Zhang, Z.}, \bibinfo{author}{Meng, D.}, \bibinfo{author}{Zhang, L.}, \bibinfo{year}{2025}.
\newblock \bibinfo{title}{{Spatial-Mamba:} {Effective} visual state space models via structure-aware state fusion}, in: \bibinfo{booktitle}{ICLR}, pp. \bibinfo{pages}{73777--73795}.
\bibitem[{Xie et~al.(2024)Xie, Cao, Xie, Khan and Pang}]{xie2024sed}
\bibinfo{author}{Xie, B.}, \bibinfo{author}{Cao, J.}, \bibinfo{author}{Xie, J.}, \bibinfo{author}{Khan, F.S.}, \bibinfo{author}{Pang, Y.}, \bibinfo{year}{2024}.
\newblock \bibinfo{title}{{SED:} {A} simple encoder-decoder for open-vocabulary semantic segmentation}, in: \bibinfo{booktitle}{CVPR}, pp. \bibinfo{pages}{3426--3436}.
\bibitem[{Xu et~al.(2025)Xu, Xu, Zhao, Han and Li}]{xu2025mamba4pass}
\bibinfo{author}{Xu, J.}, \bibinfo{author}{Xu, C.}, \bibinfo{author}{Zhao, J.}, \bibinfo{author}{Han, C.}, \bibinfo{author}{Li, H.}, \bibinfo{year}{2025}.
\newblock \bibinfo{title}{{Mamba4PASS:} {Vision} mamba for panoramic semantic segmentation}.
\newblock \bibinfo{journal}{Displays} \bibinfo{volume}{89}, \bibinfo{pages}{103058}.
\bibitem[{Yan and Xu(2026)}]{li2025pgocc}
\bibinfo{author}{Yan, C.}, \bibinfo{author}{Xu, D.}, \bibinfo{year}{2026}.
\newblock \bibinfo{title}{Progressive gaussian transformer with anisotropy-aware sampling for open vocabulary occupancy prediction}, in: \bibinfo{booktitle}{ICLR}, pp. \bibinfo{pages}{4715--4737}.
\bibitem[{Yang et~al.(2020a)Yang, Hu, Bergasa, Romera and Wang}]{yang2019pass}
\bibinfo{author}{Yang, K.}, \bibinfo{author}{Hu, X.}, \bibinfo{author}{Bergasa, L.M.}, \bibinfo{author}{Romera, E.}, \bibinfo{author}{Wang, K.}, \bibinfo{year}{2020}a.
\newblock \bibinfo{title}{{PASS:} {Panoramic} annular semantic segmentation}.
\newblock \bibinfo{journal}{IEEE Transactions on Intelligent Transportation Systems} \bibinfo{volume}{21}, \bibinfo{pages}{4171--4185}.
\bibitem[{Yang et~al.(2020b)Yang, Hu, Chen, Xiang, Wang and Stiefelhagen}]{yang2020ds}
\bibinfo{author}{Yang, K.}, \bibinfo{author}{Hu, X.}, \bibinfo{author}{Chen, H.}, \bibinfo{author}{Xiang, K.}, \bibinfo{author}{Wang, K.}, \bibinfo{author}{Stiefelhagen, R.}, \bibinfo{year}{2020}b.
\newblock \bibinfo{title}{{DS-PASS:} {Detail-sensitive} panoramic annular semantic segmentation through {SwaftNet} for surrounding sensing}, in: \bibinfo{booktitle}{IV}, pp. \bibinfo{pages}{457--464}.
\bibitem[{Yang et~al.(2026a)Yang, Qian, Wang and Yang}]{yang2026parkocc}
\bibinfo{author}{Yang, T.}, \bibinfo{author}{Qian, Y.}, \bibinfo{author}{Wang, C.}, \bibinfo{author}{Yang, M.}, \bibinfo{year}{2026}a.
\newblock \bibinfo{title}{{ParkOcc:} {A} novel dataset and benchmark for surround-view fisheye {3-D} semantic occupancy prediction in automated parking scenarios}.
\newblock \bibinfo{journal}{IEEE Transactions on Intelligent Transportation Systems} \bibinfo{volume}{27}, \bibinfo{pages}{9131--9146}.
\bibitem[{Yang et~al.(2026b)}]{yang2025daocc}
\bibinfo{author}{Yang, Z.}, et~al., \bibinfo{year}{2026}b.
\newblock \bibinfo{title}{{DAOcc:} {3D} object detection assisted multi-sensor fusion for {3D} occupancy prediction}.
\newblock \bibinfo{journal}{IEEE Transactions on Circuits and Systems for Video Technology} \bibinfo{volume}{36}, \bibinfo{pages}{1742--1753}.
\bibitem[{Yogamani et~al.(2024)Yogamani, Unger, Narayanan and Kumar}]{yogamani2024fisheyebevseg}
\bibinfo{author}{Yogamani, S.}, \bibinfo{author}{Unger, D.}, \bibinfo{author}{Narayanan, V.}, \bibinfo{author}{Kumar, V.R.}, \bibinfo{year}{2024}.
\newblock \bibinfo{title}{{FisheyeBEVSeg:} {Surround} view fisheye cameras based bird’s-eye view segmentation for autonomous driving}, in: \bibinfo{booktitle}{CVPRW}, pp. \bibinfo{pages}{1331--1334}.
\bibitem[{Yu et~al.(2025)}]{yu2024language}
\bibinfo{author}{Yu, Z.}, et~al., \bibinfo{year}{2025}.
\newblock \bibinfo{title}{Language driven occupancy prediction}, in: \bibinfo{booktitle}{ICCV}, pp. \bibinfo{pages}{7548--7558}.
\bibitem[{Yuan et~al.(2023)Yuan, Wang, Lv, Wang and Fang}]{yuan2023laformer}
\bibinfo{author}{Yuan, Z.}, \bibinfo{author}{Wang, J.}, \bibinfo{author}{Lv, Y.}, \bibinfo{author}{Wang, D.}, \bibinfo{author}{Fang, Y.}, \bibinfo{year}{2023}.
\newblock \bibinfo{title}{Laformer: Vision transformer for panoramic image semantic segmentation}.
\newblock \bibinfo{journal}{IEEE Signal Processing Letters} \bibinfo{volume}{30}, \bibinfo{pages}{1792--1796}.
\bibitem[{Zhang et~al.(2025)Zhang, Ding and Liu}]{zhang2025occloff}
\bibinfo{author}{Zhang, J.}, \bibinfo{author}{Ding, Y.}, \bibinfo{author}{Liu, Z.}, \bibinfo{year}{2025}.
\newblock \bibinfo{title}{{OccLoff:} {Learning} optimized feature fusion for {3D} occupancy prediction}, in: \bibinfo{booktitle}{WACV}, pp. \bibinfo{pages}{3096--3106}.
\bibitem[{Zhang et~al.(2022)Zhang, Yang, Ma, Rei{\ss}, Peng and Stiefelhagen}]{zhang2022bending}
\bibinfo{author}{Zhang, J.}, \bibinfo{author}{Yang, K.}, \bibinfo{author}{Ma, C.}, \bibinfo{author}{Rei{\ss}, S.}, \bibinfo{author}{Peng, K.}, \bibinfo{author}{Stiefelhagen, R.}, \bibinfo{year}{2022}.
\newblock \bibinfo{title}{Bending reality: Distortion-aware transformers for adapting to panoramic semantic segmentation}, in: \bibinfo{booktitle}{CVPR}, pp. \bibinfo{pages}{16896--16906}.
\bibitem[{Zhang et~al.(2024a)}]{zhang2024behind}
\bibinfo{author}{Zhang, J.}, et~al., \bibinfo{year}{2024}a.
\newblock \bibinfo{title}{Behind every domain there is a shift: Adapting distortion-aware vision transformers for panoramic semantic segmentation}.
\newblock \bibinfo{journal}{IEEE Transactions on Pattern Analysis and Machine Intelligence} \bibinfo{volume}{46}, \bibinfo{pages}{8549--8567}.
\bibitem[{Zhang et~al.(2024b)Zhang, Liu, Zheng and Wang}]{zhang2024goodsam}
\bibinfo{author}{Zhang, W.}, \bibinfo{author}{Liu, Y.}, \bibinfo{author}{Zheng, X.}, \bibinfo{author}{Wang, L.}, \bibinfo{year}{2024}b.
\newblock \bibinfo{title}{{GoodSAM:} {Bridging} domain and capacity gaps via segment anything model for distortion-aware panoramic semantic segmentation}, in: \bibinfo{booktitle}{CVPR}, pp. \bibinfo{pages}{28264--28273}.
\bibitem[{Zhang et~al.(2023)Zhang, Zhu and Du}]{zhang2023occformer}
\bibinfo{author}{Zhang, Y.}, \bibinfo{author}{Zhu, Z.}, \bibinfo{author}{Du, D.}, \bibinfo{year}{2023}.
\newblock \bibinfo{title}{{OccFormer:} {Dual-path} transformer for vision-based {3D} semantic occupancy prediction}, in: \bibinfo{booktitle}{ICCV}, pp. \bibinfo{pages}{9399--9409}.
\bibitem[{Zhao et~al.(2026)Zhao, Luo, Hays and Gan}]{zhao2026shelfgaussian}
\bibinfo{author}{Zhao, L.}, \bibinfo{author}{Luo, Y.}, \bibinfo{author}{Hays, J.}, \bibinfo{author}{Gan, L.}, \bibinfo{year}{2026}.
\newblock \bibinfo{title}{{ShelfGaussian:} {Shelf-supervised} open-vocabulary gaussian-based {3D} scene understanding}, in: \bibinfo{booktitle}{CVPRF}, pp. \bibinfo{pages}{1398--1409}.
\bibitem[{Zhao et~al.(2025)Zhao, Li, Shi, Imanpour and Wang}]{zhao2025dpseg}
\bibinfo{author}{Zhao, Z.}, \bibinfo{author}{Li, X.}, \bibinfo{author}{Shi, L.}, \bibinfo{author}{Imanpour, N.}, \bibinfo{author}{Wang, S.}, \bibinfo{year}{2025}.
\newblock \bibinfo{title}{{DPSeg:} {Dual-prompt} cost volume learning for open-vocabulary semantic segmentation}, in: \bibinfo{booktitle}{CVPR}, pp. \bibinfo{pages}{25346--25356}.
\bibitem[{Zheng et~al.(2024a)}]{zheng2024ops}
\bibinfo{author}{Zheng, J.}, et~al., \bibinfo{year}{2024}a.
\newblock \bibinfo{title}{Open panoramic segmentation}, in: \bibinfo{booktitle}{ECCV}, pp. \bibinfo{pages}{164--182}.
\bibitem[{Zheng et~al.(2024b)}]{zheng2024veon}
\bibinfo{author}{Zheng, J.}, et~al., \bibinfo{year}{2024}b.
\newblock \bibinfo{title}{{VEON:} {Vocabulary-enhanced} occupancy prediction}, in: \bibinfo{booktitle}{ECCV}, pp. \bibinfo{pages}{92--108}.
\bibitem[{Zheng et~al.(2026)}]{zheng2025doracamom}
\bibinfo{author}{Zheng, L.}, et~al., \bibinfo{year}{2026}.
\newblock \bibinfo{title}{{Doracamom:} {Joint} {3D} detection and occupancy prediction with multi-view {4D} radars and cameras for omnidirectional perception}.
\newblock \bibinfo{journal}{IEEE Transactions on Circuits and Systems for Video Technology} \bibinfo{volume}{36}, \bibinfo{pages}{7630--7645}.
\bibitem[{Zheng et~al.(2023a)Zheng, Pan, Luo and Wang}]{zheng2023look_neighbor}
\bibinfo{author}{Zheng, X.}, \bibinfo{author}{Pan, T.}, \bibinfo{author}{Luo, Y.}, \bibinfo{author}{Wang, L.}, \bibinfo{year}{2023}a.
\newblock \bibinfo{title}{Look at the neighbor: Distortion-aware unsupervised domain adaptation for panoramic semantic segmentation}, in: \bibinfo{booktitle}{ICCV}, pp. \bibinfo{pages}{18641--18652}.
\bibitem[{Zheng et~al.(2024c)Zheng, Zhou, Vasilakos and Wang}]{zheng2024semantics}
\bibinfo{author}{Zheng, X.}, \bibinfo{author}{Zhou, P.}, \bibinfo{author}{Vasilakos, A.V.}, \bibinfo{author}{Wang, L.}, \bibinfo{year}{2024}c.
\newblock \bibinfo{title}{Semantics distortion and style matter: Towards source-free {UDA} for panoramic segmentation}, in: \bibinfo{booktitle}{CVPR}, pp. \bibinfo{pages}{27885--27895}.
\bibitem[{Zheng et~al.(2025)Zheng, Zhou, Vasilakos and Wang}]{zheng2024360sfuda++}
\bibinfo{author}{Zheng, X.}, \bibinfo{author}{Zhou, P.Y.}, \bibinfo{author}{Vasilakos, A.V.}, \bibinfo{author}{Wang, L.}, \bibinfo{year}{2025}.
\newblock \bibinfo{title}{{360SFUDA++:} {Towards} source-free {UDA} for panoramic segmentation by learning reliable category prototypes}.
\newblock \bibinfo{journal}{IEEE Transactions on Pattern Analysis and Machine Intelligence} \bibinfo{volume}{47}, \bibinfo{pages}{1190--1204}.
\bibitem[{Zheng et~al.(2023b)Zheng, Zhu, Liu, Cao, Fu and Wang}]{zheng2023both}
\bibinfo{author}{Zheng, X.}, \bibinfo{author}{Zhu, J.}, \bibinfo{author}{Liu, Y.}, \bibinfo{author}{Cao, Z.}, \bibinfo{author}{Fu, C.}, \bibinfo{author}{Wang, L.}, \bibinfo{year}{2023}b.
\newblock \bibinfo{title}{Both style and distortion matter: Dual-path unsupervised domain adaptation for panoramic semantic segmentation}, in: \bibinfo{booktitle}{CVPR}, pp. \bibinfo{pages}{1285--1295}.
\bibitem[{Zhong et~al.(2025)}]{zhong2025omnisam}
\bibinfo{author}{Zhong, D.}, et~al., \bibinfo{year}{2025}.
\newblock \bibinfo{title}{{OmniSAM:} {Omnidirectional} segment anything model for {UDA} in panoramic semantic segmentation}, in: \bibinfo{booktitle}{ICCV}, pp. \bibinfo{pages}{23892--23901}.
\bibitem[{Zhou et~al.(2026)Zhou, Luo, Zhang, Jiang and Chen}]{zhou2026monocular_open_vocabulary}
\bibinfo{author}{Zhou, C.}, \bibinfo{author}{Luo, Y.}, \bibinfo{author}{Zhang, H.}, \bibinfo{author}{Jiang, Z.}, \bibinfo{author}{Chen, C.}, \bibinfo{year}{2026}.
\newblock \bibinfo{title}{Monocular open vocabulary occupancy prediction for indoor scenes}, in: \bibinfo{booktitle}{CVPR}, pp. \bibinfo{pages}{21627--21637}.
\bibitem[{Zhou et~al.(2025)Zhou, Wang, Wang, Wei, Dong and Yang}]{zhou2025autoocc}
\bibinfo{author}{Zhou, X.}, \bibinfo{author}{Wang, J.}, \bibinfo{author}{Wang, Y.}, \bibinfo{author}{Wei, Y.}, \bibinfo{author}{Dong, N.}, \bibinfo{author}{Yang, M.H.}, \bibinfo{year}{2025}.
\newblock \bibinfo{title}{{AutoOcc:} {Automatic} open-ended semantic occupancy annotation via vision-language guided gaussian splatting}, in: \bibinfo{booktitle}{ICCV}, pp. \bibinfo{pages}{3367--3377}.
\bibitem[{Zhu et~al.(2024)Zhu, Wang and Li}]{zhu2024nucraft}
\bibinfo{author}{Zhu, B.}, \bibinfo{author}{Wang, Z.}, \bibinfo{author}{Li, H.}, \bibinfo{year}{2024}.
\newblock \bibinfo{title}{{nuCraft:} {Crafting} high resolution {3D} semantic occupancy for unified {3D} scene understanding}, in: \bibinfo{booktitle}{ECCV}, pp. \bibinfo{pages}{125--141}.
\bibitem[{Zhu and Fan(2026)}]{zhu2026panoramic_scene_analysis}
\bibinfo{author}{Zhu, Q.}, \bibinfo{author}{Fan, L.}, \bibinfo{year}{2026}.
\newblock \bibinfo{title}{Panoramic scene analysis: A survey from distortion-aware engineering to sphere-native foundation modeling}.
\newblock \bibinfo{journal}{arXiv preprint arXiv:2606.27745} .
\bibitem[{Zuo et~al.(2025)Zuo, Zheng, Han, Yang, Pan and Lu}]{zuo2025quadricformer}
\bibinfo{author}{Zuo, S.}, \bibinfo{author}{Zheng, W.}, \bibinfo{author}{Han, X.}, \bibinfo{author}{Yang, L.}, \bibinfo{author}{Pan, Y.}, \bibinfo{author}{Lu, J.}, \bibinfo{year}{2025}.
\newblock \bibinfo{title}{{QuadricFormer:} {Scene} as superquadrics for {3D} semantic occupancy prediction}, in: \bibinfo{booktitle}{NeurIPS}, pp. \bibinfo{pages}{47779--47801}.
\bibitem[{Zuo et~al.(2023)Zuo, Zheng, Huang, Zhou and Lu}]{zuo2023pointocc}
\bibinfo{author}{Zuo, S.}, \bibinfo{author}{Zheng, W.}, \bibinfo{author}{Huang, Y.}, \bibinfo{author}{Zhou, J.}, \bibinfo{author}{Lu, J.}, \bibinfo{year}{2023}.
\newblock \bibinfo{title}{{PointOcc:} {Cylindrical} tri-perspective view for point-based {3D} semantic occupancy prediction}.
\newblock \bibinfo{journal}{arXiv preprint arXiv:2308.16896} .

\end{thebibliography}

\clearpage
\appendix
\counterwithin{figure}{section}
\counterwithin{table}{section}

In the appendix, we provide additional methodological details, including a unified interpretation of the proposed framework and the projection between Cartesian and cylindrical voxel representations. 
We then present comprehensive experimental details and analyses, covering additional qualitative results, efficiency studies, and evaluations under more challenging semantic shifts and indoor scenarios. 
Furthermore, we discuss the limitations of the proposed method and highlight potential future research directions, along with the associated social impacts.

\section{Additional Methodological Details}
\subsection{Unified and Cohesive Solution} 
\begin{figure}[pos=ht]
    \centering
    \includegraphics[width=\linewidth]{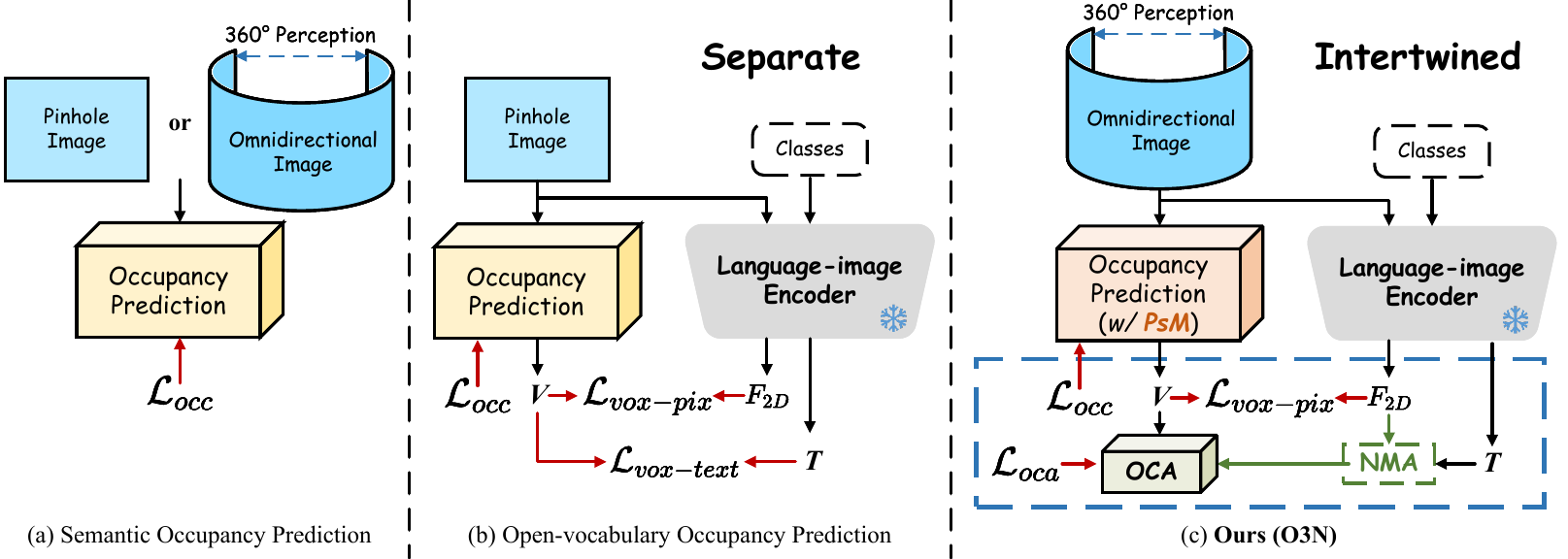}
    \caption{
    Comparison of the existing semantic occupancy (a), open-vocabulary occupancy (b) paradigms, and ours (c).
    }
    \label{fig:diffFramework}
\end{figure}

O3N builds upon the existing occupancy architecture to facilitate 3D understanding from a single omnidirectional RGB image. 
As illustrated in Fig.~\ref{fig:diffFramework}, O3N introduces three intertwined designs, namely PsM, OCA, and NMA, which jointly address panoramic spatial modeling, semantic generalization, and cross-modal representation alignment, respectively. 
Regular Cartesian grids misalign with the 360{\textdegree} field of view without PsM, causing spatial misregistration that corrupts both geometry and semantics; OCA builds upon PsM to ensure geometric-semantic consistency in voxel space, enabling semantic propagation across entire 3D regions, whereas existing methods rely on discrete closed-set supervision; finally, integrating NMA closes the loop, which constructs a robust \textit{``pixel-voxel-text''} representational triplet. Without the alignment, backpropagating through the text encoder forces the model to overfit to training semantics, impairing generalization to unseen concepts.

\subsection{Cubic \textit{vs.} Cylindrical Voxel Projections}
\begin{figure}[pos=!t]
    \centering
    \includegraphics[width=.8\linewidth]{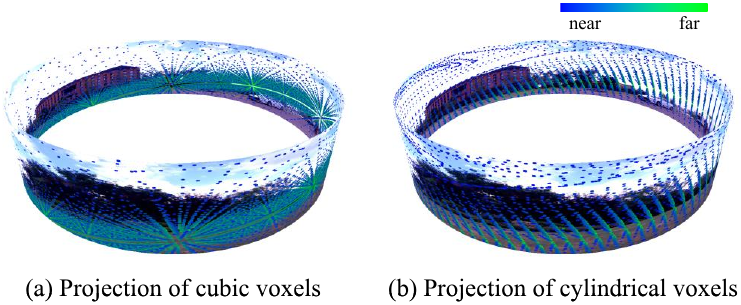}
    \caption{
    \textbf{Cubic \textit{vs.} cylindrical voxel projections on an omnidirectional image.} The projection of Cartesian cubic voxels in the omnidirectional image reflects the sparsity and imbalance of projection points.
    }
    \label{fig:cubic_vs_cyclindrical}
\end{figure}

We visualize the projection patterns induced by different voxel parameterizations on an equirectangular omnidirectional image from QuadOcc. As shown in Fig.~\ref{fig:cubic_vs_cyclindrical}, each point represents a projected voxel center, while its color encodes the normalized radial depth of the corresponding occupied voxel.
For an axis-aligned Cartesian cubic voxel grid, uniform metric spacing in 3D space becomes non-uniform angular sampling after projection onto the equirectangular image.
The projected voxel centers consequently exhibit fan-like patterns. 
Moreover, as the radial distance increases, a fixed vertical displacement subtends a progressively smaller elevation angle. Consequently, distant road surfaces, vehicles, pedestrians, and roadside structures may be concentrated near the equatorial region of the panorama, weakening the visual evidence available for voxel-level discrimination.
Therefore, although the Cartesian representation preserves a regular metric structure in 3D space, it does not explicitly conform to the radial organization and azimuthal periodicity of omnidirectional observations.

By contrast, a camera-centric omnidirectional cylindrical voxel grid parameterized by radius, azimuth, and height provides a uniform discretization along the azimuthal dimension, which naturally corresponds to the horizontal coordinate of the equirectangular image.
For fixed radial and height bins, varying the azimuth produces horizontally organized projection patterns, thereby better preserving the angular ordering of surrounding structures.
Nevertheless, the metric extent of each angular bin increases with radial distance, making the cylindrical representation less suitable for maintaining uniform Cartesian geometry.
These complementary properties motivate the joint use of cylindrical and Cartesian voxel representations in PsM.
The cylindrical branch facilitates camera-centric azimuthal modeling, whereas the Cartesian branch retains the regular metric geometry required by downstream occupancy-based navigation and spatial reasoning.

\section{More Results}
\subsection{More Qualitative Results}
\begin{figure*}[pos=!t]
    \centering
    \includegraphics[width=\textwidth]{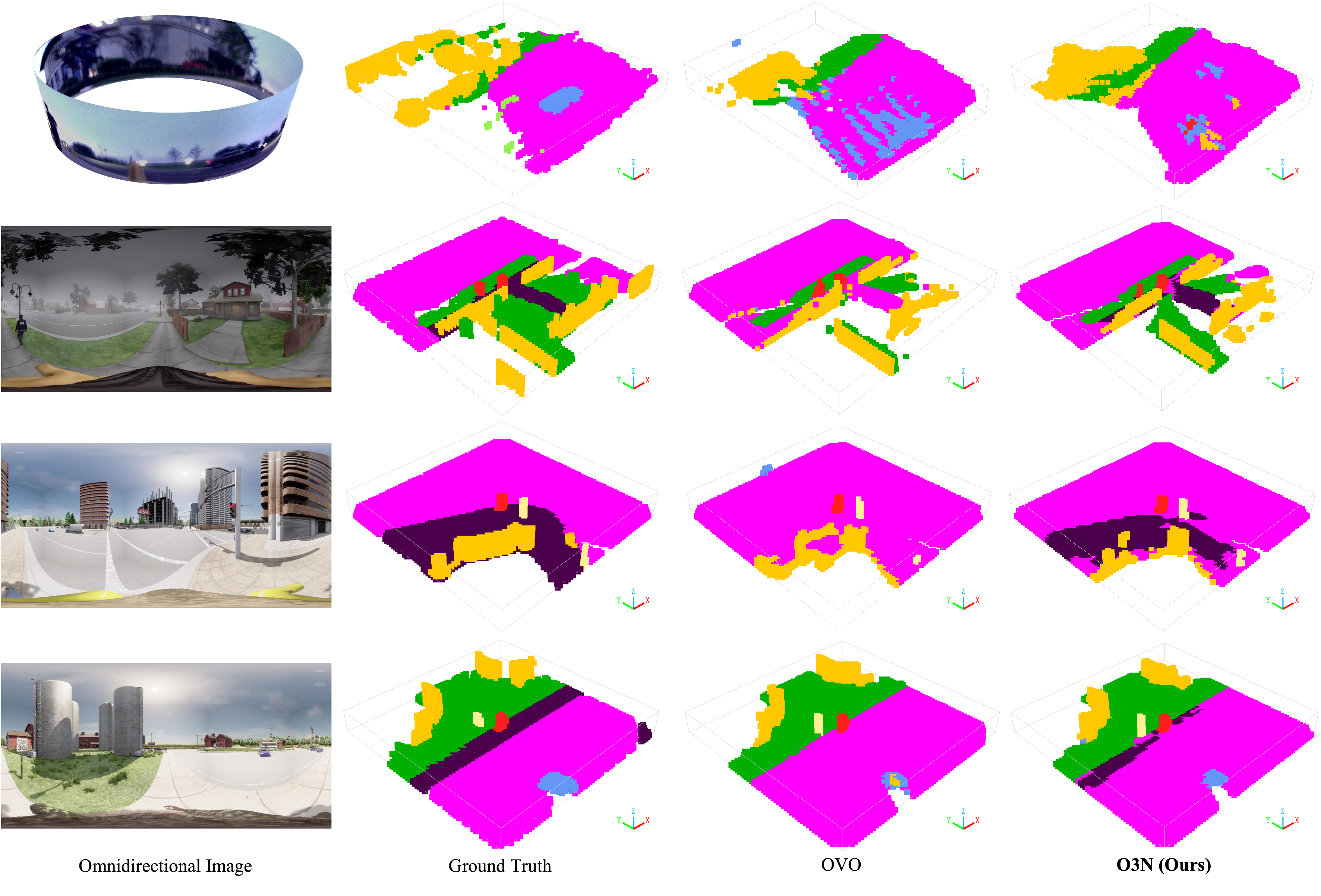}
    \caption{
    \textbf{More visualization results on QuadOcc~\citep{shi2026oneocc} and Human360Occ~\citep{shi2026oneocc} datasets.}
    }
    \label{fig:visualResults_sup}
\end{figure*}

In Fig.~\ref{fig:visualResults_sup}, we provide more qualitative results for open-vocabulary occupancy prediction on QuadOcc~\citep{shi2026oneocc} and Human360Occ~\citep{shi2026oneocc} datasets. The predictions generated by O3N exhibit distinctly clearer geometric structures and improved semantic discrimination, particularly for unseen object categories and the overall scene layout. 
By contrast, OVO~\citep{tan2023ovo} tends to generate fragmented occupancy structures and less discriminative semantic predictions in challenging regions. 
These qualitative observations are consistent with the quantitative improvements in overall and \emph{Novel} mIoU, further demonstrating the effectiveness of O3N in omnidirectional open-vocabulary occupancy prediction.

\subsection{Efficiency Analysis}
As shown in Tab.~\ref{tab:efficiency}, our O3N framework is highly competitive in terms of both inference speed and memory consumption. Compared to the baseline OVO~\citep{tan2023ovo}, our efficient variant, O3N (SGN-S), achieves higher accuracy ($+1.19$ mIoU) while being substantially faster ($13.36$ vs. $10.67$ FPS) and more memory-efficient ($3.52$ vs. $4.28$ GB). Even when compared to fully supervised methods like VoxFormer-S~\citep{li2023voxformer} and MonoScene~\citep{cao2022monoscene}, O3N maintains a favorable balance between performance and computational cost. While OneOcc~\citep{shi2026oneocc} is optimized for closed-set tasks, O3N provides the essential capability of novel class recognition with comparable real-time performance. These results demonstrate that O3N provides a superior efficiency-performance trade-off for omnidirectional open-vocabulary occupancy prediction.

\begin{table}[pos=!t]
    \centering
    \caption{Efficiency comparison on QuadOcc~\citep{shi2026oneocc}.}
    \label{tab:efficiency}
    \resizebox{\linewidth}{!}{%
    \begin{tabular}{l|c|cc}
        \toprule
        Method & mIoU & \textbf{FPS} $\uparrow$ & \textbf{Mem. (GB)} $\downarrow$ \\
        \midrule
        VoxFormer-S~\citep{li2023voxformer} & 9.36 & 7.58 & 4.84 \\
        MonoScene~\citep{cao2022monoscene} & \underline{19.19} & \underline{8.29} & \underline{2.60} \\
        OneOcc~\citep{shi2026oneocc} & \textbf{20.56} & \textbf{14.30} & \textbf{1.82} \\
        \midrule
        OVO~\citep{tan2023ovo} & 14.33 & \underline{10.67} & \underline{4.28} \\
        \textbf{Ours (SGN-S)} & \underline{15.52} & \textbf{13.36} & \textbf{3.52} \\
        \textbf{Ours} & \textbf{16.54} & 9.41 & 4.97 \\
        \bottomrule
    \end{tabular}
    }
\end{table}

\begin{table*}[pos=!t]
\centering
\caption{\textbf{Omnidirectional 3D occupancy prediction results on Human360Occ~\citep{shi2026oneocc} Heter (cross-city) split.} C: Camera.}
\label{tab:comparison_h3o_heter}
\resizebox{\textwidth}{!}{
\begin{tabular}{l|c|c|cccccccc|cccc}
    \toprule
    & & & \multicolumn{8}{c}{\textbf{Novel Class}}& \multicolumn{4}{c}{\textbf{Base Class}}\\
    \textbf{Method}& \textbf{Input}& \textbf{mIoU}
    & \ColHeadHeter{hthreeoRoad}{road}{road} 
    & \ColHeadHeter{hthreeoSidewalk}{sidewalk}{sidewalk} 
    & \ColHeadHeter{hthreeoBuilding}{building}{building} 
    & \ColHeadHeter{hthreeoCar}{car}{car} 
    & \ColHeadHeter{hthreeoTruck}{truck}{truck} 
    & \ColHeadHeter{hthreeoBus}{bus}{bus} 
    & \ColHeadHeter{hthreeoTwoWheeler}{two\_whe.}{twoWheeler} 
    & mean 
    & \ColHeadHeter{hthreeoVegetation}{veg.}{vegetation} 
    & \ColHeadHeter{hthreeoPerson}{person}{person} 
    & \ColHeadHeter{hthreeoPole}{pole}{pole} 
    & mean \\
    \midrule
    \hline
    \rowcolor{mygray} \multicolumn{15}{c}{\emph{\textbf{Fully-supervised}}}\\
    \midrule
    VoxFormer-S~\citep{li2023voxformer}& \emph{C}& 10.63& 19.02& 15.82& 2.84& 0.95& 0.10& 0.00& 0.01& 5.53& 5.86& 61.62& 0.10& 22.53\\
    OccFormer~\citep{zhang2023occformer}& \emph{C}& 20.87& 47.00& 39.01& 10.14& 11.84& 1.52& 0.05& 0.77& 15.76& 24.85& 66.60& 6.93& 32.79\\
    SGN-S~\citep{mei2024sgn}& \emph{C}& 20.60& 43.47& 24.70& 9.19& 17.21& 8.21& 1.85& 5.08& 15.67& 20.27& 61.10& 14.89& 32.09\\
    SGN-T~\citep{mei2024sgn}& \emph{C}& 20.02& 42.57& 23.94& 9.42& 18.22& 10.21& 1.55& 3.11& 15.57& 15.84& 63.44& 11.93& 30.40\\
    MonoScene~\citep{cao2022monoscene}& \emph{C}& 24.15& 49.68& 41.41& 11.04& 20.34& 10.36& 1.98& 2.79& 19.66& 20.57& 69.68& 13.69& 34.65\\
    OneOcc~\citep{shi2026oneocc}& \emph{C}& 32.23& 58.60& 48.28& 16.34& 30.68& 19.35& 12.87& 14.16& 28.61& 25.98& 72.89& 23.11& 40.66\\
    \hline
    \rowcolor{mygray} \multicolumn{15}{c}{\emph{\textbf{Open-vocabulary}}}\\
    \midrule
    OVO~\citep{tan2023ovo} & \emph{C}& 18.29& 29.89& 9.99& 12.36& 15.09& 2.01& 0.23& 0.00& 9.94& 22.23& 68.30& 22.85& 37.79\\
    {\cellcolor{mygreen!10}}\textbf{Ours} & 
    {\cellcolor{mygreen!10}}\emph{C}&
    {\cellcolor{mygreen!10}}18.98&
    {\cellcolor{mygreen!10}}31.79&
    {\cellcolor{mygreen!10}}15.61&
    {\cellcolor{mygreen!10}}10.59&
    {\cellcolor{mygreen!10}}16.46&
    {\cellcolor{mygreen!10}}0.00&
    {\cellcolor{mygreen!10}}0.00&
    {\cellcolor{mygreen!10}}0.00&
    {\cellcolor{mygreen!10}}10.64&
    {\cellcolor{mygreen!10}}22.79&
    {\cellcolor{mygreen!10}}69.63&
    {\cellcolor{mygreen!10}}22.89&
    {\cellcolor{mygreen!10}}38.44\\
    \bottomrule
\end{tabular}
}
\end{table*}

\begin{table*}[pos=!t]
\centering
\caption{\textbf{3D occupancy prediction results on NYUv2~\citep{silberman2012indoor}}. C: Camera.}
\label{tab:comparison_nyuv2}
\begin{threeparttable}
\resizebox{\textwidth}{!}{
\begin{tabular}{l|c|c|cccc|ccccccccc}
    \toprule
    & & & \multicolumn{4}{c}{\textbf{Novel Class}}& \multicolumn{9}{c}{\textbf{Base Class}}\\
    \textbf{Method}& \textbf{Input}& \textbf{mIoU} 
    & \ColHeadNyu{bed}{bed}{bed} 
    & \ColHeadNyu{table}{table}{table} 
    & \ColHeadNyu{other}{other}{other} 
    & mean
    & \ColHeadNyu{ceiling}{ceiling}{ceiling} 
    & \ColHeadNyu{floor}{floor}{floor} 
    & \ColHeadNyu{wall}{wall}{wall} 
    & \ColHeadNyu{window}{window}{window} 
    & \ColHeadNyu{chair}{chair}{chair} 
    & \ColHeadNyu{sofa}{sofa}{sofa} 
    & \ColHeadNyu{tvs}{tvs}{tvs} 
    & \ColHeadNyu{furniture}{furniture}{furniture} 
    & mean \\
    \midrule
    \hline
    \rowcolor{mygray} \multicolumn{16}{c}{\emph{\textbf{Fully-supervised}}}\\
    \midrule
    LMSCNet$^\text{rgb}$~\citep{roldao2020lmscnet}& \emph{C}& 15.88& 32.03& 6.57& 4.39& 14.33& 4.49& 88.41& 4.63& 0.25& 3.94& 15.44& 0.02& 14.51& 16.46\\
    SGN-S~\citep{mei2024sgn}& \emph{C}& 27.71& 48.59& 14.08& 12.07& 24.91& 14.67& 93.56& 13.36& 10.35& 14.64& 37.47& 15.75& 30.31& 28.76\\
    MonoScene~\citep{cao2022monoscene}& \emph{C}& 26.94& 48.19& 15.13& 12.94& 25.42& 8.89& 93.50& 12.06& 12.57& 13.72& 36.11& 15.22& 27.96& 27.50\\
    \hline
    \rowcolor{mygray} \multicolumn{16}{c}{\emph{\textbf{Open-vocabulary}}}\\
    \midrule
    OVO$^{*}$~\citep{tan2023ovo}& \emph{C}& 21.13& 29.37& 3.91& 8.34& 13.88& 6.91& 92.47& 8.52& 7.72& 9.92& 31.97& 8.54& 24.77& 23.85\\
    {\cellcolor{mygreen!10}}\textbf{Ours}& 
    {\cellcolor{mygreen!10}}\emph{C}& 
    {\cellcolor{mygreen!10}}23.21&
    {\cellcolor{mygreen!10}}31.92&
    {\cellcolor{mygreen!10}}8.01&
    {\cellcolor{mygreen!10}}6.14&
    {\cellcolor{mygreen!10}}15.36&
    {\cellcolor{mygreen!10}}10.48&
    {\cellcolor{mygreen!10}}93.57&
    {\cellcolor{mygreen!10}}9.41&
    {\cellcolor{mygreen!10}}10.32&
    {\cellcolor{mygreen!10}}12.29&
    {\cellcolor{mygreen!10}}32.86&
    {\cellcolor{mygreen!10}}15.03&
    {\cellcolor{mygreen!10}}25.26& 
    {\cellcolor{mygreen!10}}26.15\\
    \bottomrule
\end{tabular}
}
\begin{tablenotes}[]
\item[$*$] denotes results reproduced using the official code implementation.
\end{tablenotes}
\end{threeparttable}
\end{table*}

\subsection{Generalization under More Challenging Semantic Shifts and Indoor Scenarios}
We extend our evaluation to investigate the robustness of O3N under more extreme semantic distribution shifts and its applicability to indoor environments. While standard benchmarks provide a solid foundation, understanding model behavior under significant domain gaps is crucial for real-world deployment. 
To evaluate performance under severe semantic distribution shifts, we supplement our evaluation with experiments on H3O-Heter, a \emph{cross-city} split with extreme intra-category visual-semantic variations, as shown in Table~\ref{tab:comparison_h3o_heter}. 
O3N demonstrates robustness under significant semantic distribution shifts, outperforming OVO by $0.7\%$ in \emph{novel} mIoU and $0.69\%$ in overall mIoU. 
We further evaluate our method on the indoor pinhole dataset NYUv2~\citep{silberman2012indoor}. As shown in Table~\ref{tab:comparison_nyuv2}, it consistently achieves superior performance, particularly on unseen semantics ($+1.48\%$).

\section{Discussions}
\subsection{Limitations and Future Work}
The performance of omnidirectional open-vocabulary occupancy prediction is dependent on the vision-language models, but a vision-language model tailored for omnidirectional embodied perception is scarce in the literature. 
The extracted open-vocabulary 2D image pixel features are prone to semantic shifts, limited generalization, and distribution bias in valid voxel sampling. 
Furthermore, the non-uniform pixel-to-voxel correspondences inherent in omnidirectional projection may introduce additional biases in voxel-level semantic representations, further complicating cross-modal alignment.
Another limitation concerns the recognition of rare novel categories under highly imbalanced semantic distributions. While structured spatial modeling and cost aggregation encourage geometric–semantic consistency and help suppress spatially extensive erroneous predictions, they may also limit the prediction coverage of rare categories with insufficient visual evidence. This suggests a potential trade-off between false-positive suppression and rare-category recall, which remains challenging in omnidirectional open-vocabulary occupancy prediction.

Future research could explore self-supervised learning methods for omnidirectional visual scenes to enhance generalizability in semantic understanding. Geometry-aware feature projection could further improve the reliability of pixel-to-voxel semantic correspondences. In addition, uncertainty-aware semantic matching and adaptive cost aggregation represent promising directions for improving sensitivity to rare categories while maintaining geometric-semantic consistency.

\begin{figure*}[pos=!t]
    \centering
    \includegraphics[width=\textwidth]{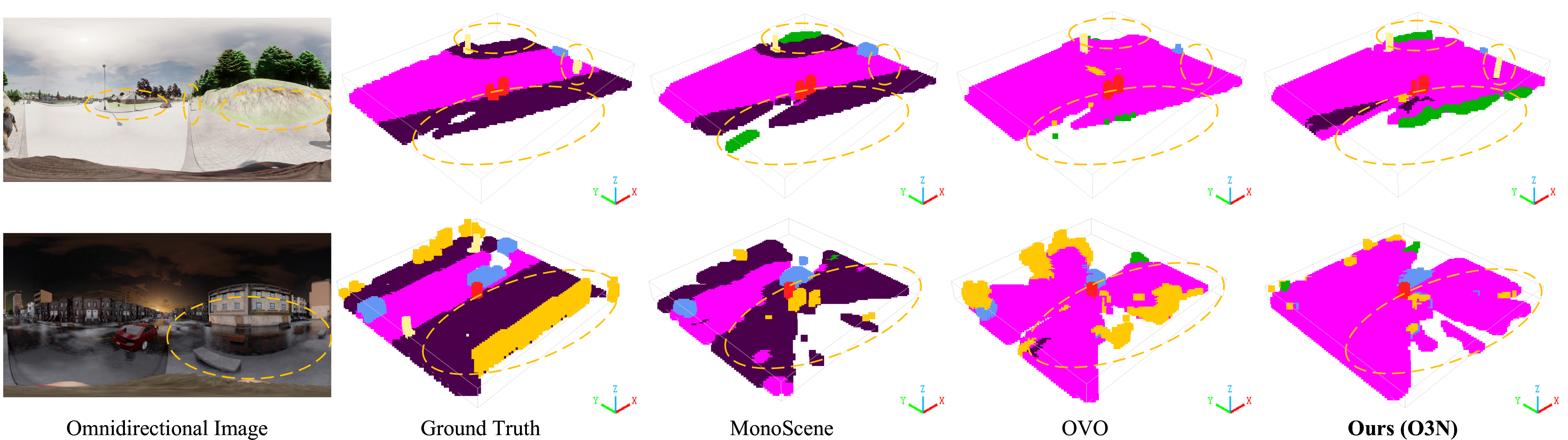}
    \caption{
    \textbf{Failure case analysis on H3O~\citep{shi2026oneocc} Homo (within-city).} Similar to MonoScene~\citep{cao2022monoscene}, O3N employs the same 2D-to-3D projection strategy (FLoSP), resulting in limited sensitivity to spatial depth. This limitation causes inaccuracies in accurately perceiving the full spatial extent of the scene.
    }
    \label{fig:failure_case}
\end{figure*}

\subsection{Failure Case Analysis}
Fig.~\ref{fig:failure_case} presents two typical failure cases of O3N. 
Depth ambiguity is an inherent challenge in monocular 3D reconstruction. While depth-based Lift-Splat-Shoot (LSS) projection~\citep{philion2020lift} is common in pinhole vision, panoramic depth estimation contains substantial noise in the predicted depth maps, which severely compromises spatial perception accuracy. Given these challenges, we adopt the Features Line of Sight 2D-to-3D Projection (FLoSP) projection strategy~\citep{cao2022monoscene,mei2024sgn}, verified in OneOcc~\citep{shi2026oneocc}, as it is more concise and robust than learnable LSS~\citep{philion2020lift} and the cross-attention mechanism~\citep{li2024bevformer}, and optimal for panoramic settings. 
However, the FLoSP method is insensitive to overall spatial depth, so in the first case, nearly all methods using FLoSP exhibit misalignment in spatial range perception. Specifically, when the sidewalk is adjacent to vegetation, all methods demonstrate additional prediction errors. Furthermore, due to the visual similarity between the pole and sidewalk, both MonoScene and OVO fail to detect the pole, while O3N accurately reconstructs the occupancy state of the pole. This demonstrates that O3N can effectively perceive both geometric and semantic information in space, thereby showcasing enhanced generalization ability.

In the second case, in a rainy street scene at dusk, due to the dim lighting and interference from rain-reflected light, MonoScene misidentifies most of the sidewalk as unoccupied (\emph{empty}). OVO and O3N also encounter similar issues, but, compared to OVO, O3N more robustly reconstructs the occupancy state of the entire scene and demonstrates clearer semantic awareness. 
For example, OVO reconstructs the \emph{sidewalk} as a \emph{building}, while O3N reconstructs it as \emph{road}, which is more semantically similar to the \emph{sidewalk}.

These failures demonstrate that while current 2D-to-3D projection strategies can effectively extend omnidirectional images into 3D space, they still have inherent limitations in spatial localization and are prone to structural misjudgments. Therefore, enhancing the spatial depth perception of models to mitigate projection-induced geometric biases constitutes an important direction for future research. Furthermore, to address complex environmental factors, such as insufficient lighting and reflections caused by rain or snow, large-scale data augmentation and domain adaptation across different weather conditions and scenes can improve the robustness and generalization performance, especially in extreme weather and previously unseen scenarios.

\subsection{Societal Impacts}
Omnidirectional open-vocabulary occupancy prediction has the potential to enhance the perceptual intelligence and safety of intelligent systems, while supporting future embodied intelligent robotics applications in healthcare, smart services, and human-centered environments. However, the method remains imperfect in prediction accuracy, and practical deployment requires careful consideration of uncertainty, robustness, and potential risks.

\end{document}